%% file: iclr2027_conference.tex
\pdfoutput=1
\documentclass{article}
\usepackage{iclr2027_conference,times}

\input{math_commands.tex}

\usepackage{hyperref}
\usepackage{url}
\usepackage{amsmath,amssymb}
\usepackage{amsthm}
\usepackage{algorithm}
\usepackage[noend]{algpseudocode}
\newtheorem{proposition}{Proposition}
\newtheorem{lemma}{Lemma}
\newtheorem{corollary}{Corollary}
\theoremstyle{remark}
\newtheorem{remark}{Remark}
\theoremstyle{plain}
\usepackage{booktabs}
\usepackage{graphicx}
\usepackage[table]{xcolor}
\usepackage{tikz}
\usepackage{paralist}
\usepackage{pgfplots}
\pgfplotsset{compat=1.18}
\usepgfplotslibrary{groupplots}
\usetikzlibrary{arrows.meta, positioning, fit, backgrounds, plotmarks}

\definecolor{cbase}{HTML}{999999}
\definecolor{cthicket}{HTML}{0072B2}
\definecolor{cttmv}{HTML}{E69F00}
\definecolor{clogit}{HTML}{D55E00}
\definecolor{csoup}{HTML}{009E73}
\definecolor{cextra}{HTML}{CC79A7}
\definecolor{cwrite}{HTML}{009E73}
\definecolor{cmember}{HTML}{F0E442}

\usepackage[skip=3pt]{caption}
\makeatletter
\newcommand{\maketitlesupplementary}{%
  \begin{center}
    {\LARGE \bf \papertitle \par}
    \vskip 0.5em
    {\large Supplementary Material \par}
    \vskip 1em
    \vskip 1em
  \end{center}
}
\makeatother

\pgfplotsset{
  paperaxis/.style={
    tick align=outside, tick pos=left,
    every axis plot/.append style={line width=0.9pt},
    label style={font=\small}, tick label style={font=\footnotesize},
    title style={font=\small\bfseries, yshift=-1ex},
    legend cell align=left,
    legend style={font=\footnotesize, draw=none, fill=none, inner sep=1pt},
    grid=major, grid style={gray!18, line width=0.3pt},
    axis line style={gray!55},
  },
  every axis/.append style={line width=0.5pt},
}

\newcommand{\papertitle}{Rethinking Neural Thickets Without Weight Perturbations}
\title{\papertitle}

\author{%
Nicolas Dufour$^{1}$ \quad Elouan Gard\`es$^{2,3}$ \quad Amil Dravid$^{4}$ \quad Patrick P\'erez$^{5}$\thanks{Work done while at Kyutai.} \quad Loic Landrieu$^{3}$ \\
$^{1}$Kyutai \quad $^{2}$Meta FAIR \quad $^{3}$LIGM, CNRS, Univ Gustave Eiffel, ENPC, IP Paris \\
$^{4}$UC Berkeley \quad $^{5}$Google DeepMind \\
\texttt{nicolas.dufour@kyutai.org}
}

\iclrfinalcopy
\begin{document}

\maketitle

\begin{abstract}
The Neural Thickets hypothesis posits that pretrained language models lie in dense neighborhoods of task-specific experts in weight space, which random Gaussian perturbations can uncover. We reproduce the reported gains, but show that they do not require this geometric explanation. Perturbing a single decoder block already recovers much of the improvement, while keeping the model entirely frozen and adding a random bias to its output logits matches or exceeds full-weight perturbations on most reasoning tasks. This suggests an alternative explanation based on output diversity, validation-based selection, and voting. We formalize this view as \emph{Perturb-Select-Vote} (PSV), a general procedure that generates perturbed predictors, selects high-performing members on a small validation set, and aggregates their predictions.

This interpretation connects Neural Thickets to established inference-time methods, including sampling, self-consistency, and verifier-guided selection. A first-order analysis further shows that small weight perturbations act primarily through their induced logit shifts. We extend this analysis to vision, where logit PSV substantially improves performance under domain shift, including on Earth observation, while full-weight perturbations provide little additional benefit. Together, these results suggest that the gains attributed to nearby weight-space experts can instead be obtained through output diversity, selection, and voting. Logit PSV realizes these benefits with a fully frozen model, providing a simple, training-free alternative to full-weight perturbation ensembles across language and vision. Project page can be found here: \url{https://kyutai.org/logit-psv}.
\end{abstract}

\input{sections/1_introduction}

\input{sections/2_related_work}

\input{sections/3_preliminaries}

\input{sections/6_minimal_method}

\input{sections/77_vision}

\input{sections/7_discussion}

\input{sections/9_statements}

\bibliography{iclr2027_conference}
\bibliographystyle{iclr2027_conference}

\clearpage
\maketitlesupplementary
\appendix
\input{sections/8_appendix}

\end{document}

%% file: math_commands.tex
\usepackage{amsmath,amsfonts,bm}

\def\eqref#1{equation~\ref{#1}}

\def\1{\bm{1}}

\def\vtheta{{\bm{\theta}}}

\DeclareMathAlphabet{\mathsfit}{\encodingdefault}{\sfdefault}{m}{sl}
\SetMathAlphabet{\mathsfit}{bold}{\encodingdefault}{\sfdefault}{bx}{n}

\newcommand{\taskpoint}[4]{%
  \addplot[
    only marks,
    mark size=1.65pt,
    #3,
    #4,
    forget plot
  ] coordinates {(#1,#2)};
}%
\newcommand{\legendmark}[2][]{%
  \tikz[baseline=-4.5ex]
    \draw[
      only marks,
      mark size=1.6pt,
      #1,
      #2
    ] plot coordinates {(0,0)};
}

%% file: sections/1_introduction.tex
\section{Introduction}
\label{sec:intro}

\citet{gan2026thickets} reported a striking phenomenon: randomly perturbing the weights of a pretrained language model can produce many variants that outperform it on specific reasoning tasks. The empirical gains are substantial, reaching up to {$52.8$ accuracy points (Countdown with Llama-3.1-8B-Instruct, from $10.8\%$ to $63.6\%$)}, and require little compute compared to fine-tuning. The authors propose a geometric explanation: pretrained weights of sufficiently large models are surrounded by a dense \emph{thicket} of task-specialized models, recoverable by blind Gaussian perturbations.

This paper first asks whether the gain requires perturbing the entire network. It does not: perturbing only a small fraction of the model, or even a single decoder block, recovers much of the gain. Pushing this observation to its limit, weight perturbations can be removed altogether: the model is kept completely frozen and only its output is perturbed: each member differs by a random Gaussian bias, fixed throughout autoregressive generation, added to the logits before greedy decoding. This logit-only variant performs no additional computation inside the network and modifies no model weight. Nevertheless, it recovers, and {in some settings} improves on, the gains of full-weight perturbations. On OLMo-3-7B-Instruct, it raises MATH-500 accuracy from $63.3\%$ to {$81.3\%$}, compared with $74.7\%$ for the full-weight perturbation ensemble.%

\begin{figure}[t]
\centering
\input{figures/fig_psv_compare}
\caption{{\textbf{Full-weight PSV versus logit PSV.}
Both methods share the same selection and voting stages (\textit{right}), but differ in the perturbation (\textit{left}). Neural Thickets (\textit{top}) perturb the full $d_{\vtheta}$-dimensional parameter vector, producing $N$ distinct networks. Logit PSV (\textit{bottom}) keeps one frozen network and adds a $V$-dimensional bias to the output logits, fixed across decoding steps {and inputs}.}}
\label{fig:pipeline}
\end{figure}

The Neural Thickets interpretation attributes these gains to local experts in weight space. Our experiments suggest a simpler account: perturbations induce output diversity, validation-based selection retains useful members, and voting aggregates their predictions. Neural Thickets can therefore be viewed as an instance of a broader \emph{perturb-select-vote} (PSV) template: sample random perturbations, score the resulting members on a small validation set, retain the top $K$, and vote at test time.
This output-space view connects PSV to standard inference-time methods. A logit bias modifies the relative probability of each token, and greedy decoding under random biases defines a perturb-and-MAP sampler~\citep{papandreou2011perturb}. Selecting members by validation performance retains the biases that perform best on the target task, analogously to best-of-$N$ and verifier-guided selection~\citep{cobbe2021verifiers}, while voting relates PSV to self-consistency~\citep{wang2023selfconsistency}. A first-order analysis further supports this view: small weight perturbations act, to leading order, through the logit shifts they induce. This paper then isolates the roles of sampling, selection, and voting: on strong base models, temperature-tuned sampling with majority voting already matches full-weight PSV, showing that output diversity and ensembling explain most of the gain. On weaker models, validation-based selection remains important as it filters unreliable {members} before aggregation.

{Finally, the analysis extends to pretrained vision models under domain shift. In this setting, the analogue of a task expert is a domain expert: a perturbed representation that improves linear-probe performance on a target domain. Full-weight PSV, which perturbs the vision backbone and fits one probe per member, is compared to logit-only PSV, which fits a single probe and perturbs only the resulting class logits. The results support the same conclusion as in language: perturbing the logits is sufficient to recover the gains of full-weight perturbations, and often improves out-of-domain performance while leaving the backbone unchanged.}

Our results do not rule out useful structure in the neighborhood of pretrained weights. They show, however, that such structure is not required to obtain the observed gains. PSV is better understood as output diversification, validation-based filtering, and ensembling. Beyond its role in interpreting Neural Thickets, PSV yields a simple, training-free output-space procedure that substantially improves performance on specialized tasks and domains.

\paragraph{Contributions.}
\begin{compactitem}
    \item \textbf{Full-weight perturbations are unnecessary.} Perturbing even a single decoder block recovers much of the gain of full-weight perturbations.
    \item \textbf{Logit PSV.} Applying PSV directly to the output logits of a fully frozen model matches {full-weight perturbation ensembles on reasoning tasks, and sometimes exceeds them}.
    \item \textbf{An output-space interpretation of PSV.} Perturbations induce output diversity, selection retains high-reward members, voting aggregates their predictions, and a first-order analysis connects weight perturbations to their induced logit shifts.
    \item \textbf{Extension to vision.} Under domain shift, output-space perturbations recover the gains of full-weight perturbations for pretrained vision backbones as well.
\end{compactitem}

%% file: figures/fig_psv_compare.tex
\resizebox{\linewidth}{!}{%
\begin{tikzpicture}[
    font=\small,
    >=latex,
    flow/.style={->, line width=0.9pt, draw=black!55},
    band/.style={rounded corners=4pt, draw=black!20, minimum width=10.2cm,
        minimum height=2.85cm, inner sep=0pt, anchor=north west},
    plus/.style={draw=black, thick, circle, inner sep=0pt, minimum size=0.42cm,
        font=\normalsize},
    noise/.style={draw=black!25, rounded corners=2pt, fill=white, inner sep=2.5pt,
        font=\footnotesize},
    note/.style={font=\footnotesize, align=left, anchor=west},
]

\newcommand{\cnet}[5]{%
\begin{scope}[shift={(#1,#2)}, scale=#3]
    \foreach \y in {-0.45,0,0.45}{\fill[#4] (0,\y) circle (0.06);}
    \foreach \y in {-0.6,-0.2,0.2,0.6}{\fill[#4] (0.5,\y) circle (0.06);}
    \foreach \y in {-0.25,0.25}{\fill[#4] (1.0,\y) circle (0.06);}
    \foreach \a in {-0.45,0,0.45}{\foreach \b in {-0.6,-0.2,0.2,0.6}{
        \draw[#4, opacity=.45, line width=.6pt] (0,\a)--(0.5,\b);}}
    \foreach \a in {-0.6,-0.2,0.2,0.6}{\foreach \b in {-0.25,0.25}{
        \draw[#4, opacity=.45, line width=.6pt] (0.5,\a)--(1.0,\b);}}
    \coordinate (#5-w) at (-0.12,0);
    \coordinate (#5-e) at (1.12,0);
    \coordinate (#5-s) at (0.5,-0.72);
    \coordinate (#5-n) at (0.5,0.72);
\end{scope}}

\newcommand{\ccard}[4]{%
    \filldraw[draw=black!20, fill=white, rounded corners=2pt]
        (#1,{#2-#4/2}) rectangle ({#1+#3},{#2+#4/2});}

\newcommand{\chist}[9]{%
\pgfmathsetmacro{\hmax}{max(#4,#5,#6,#7,#8)}
\begin{scope}[shift={(#1,#2)}]
    \draw[black!30] (0,-0.32) -- (0.86,-0.32);
    \foreach \h [count=\k] in {#4,#5,#6,#7,#8}{
        \pgfmathsetmacro{\isarg}{\h==\hmax ? 1 : 0.35}
        \fill[#3, fill opacity=\isarg] ({0.04+(\k-1)*0.16},-0.32) rectangle ++(0.12,\h);}
    \coordinate (#9-w) at (-0.08,0);
    \coordinate (#9-e) at (0.94,0);
    \coordinate (#9-s) at (0.43,-0.42);
    \coordinate (#9-n) at (0.43,0.42);
\end{scope}}

\newcommand{\clock}[2]{%
\begin{scope}[shift={(#1,#2)}]
    \draw[black!60, line width=0.8pt] (-0.07,0.02) -- (-0.07,0.09)
        arc (180:0:0.07) -- (0.07,0.02);
    \fill[black!60, rounded corners=0.5pt] (-0.11,-0.12) rectangle (0.11,0.03);
\end{scope}}

\def\yA{0}
\def\yB{-3.1}
\def\xh{5.25}

\node[band, fill=cthicket!5] (bandA) at (0,{\yA+1.55}) {};
\node[band, fill=clogit!6]   (bandB) at (0,{\yB+1.55}) {};

\node[font=\bfseries, text=cthicket!80!black, anchor=north west]
    at ([xshift=2mm, yshift=-1.5mm]bandA.north west)
    {Full-weight PSV  (neural thickets)};
\node[font=\bfseries, text=clogit!80!black, anchor=north west]
    at ([xshift=2mm, yshift=-1.5mm]bandB.north west)
    {Logit PSV (ours)};

\cnet{0.45}{\yA-0.15}{0.75}{black!70}{baseA}
\node[font=\footnotesize, anchor=north] at (baseA-s) {$f_{\vtheta}$};

\node[plus] (plusA) at (2.05,\yA-0.15) {$+$};
\node[noise] (epsA) at (2.05,\yA+0.64) {$\sigma\boldsymbol{\epsilon}_i \in \mathbb{R}^{d_{\vtheta}}$};
\node[font=\scriptsize, text=black!55, anchor=north] at (plusA.south) {into weights};

\foreach \dx/\dy/\col [count=\j] in
    {0.30/0.24/csoup!85!black, 0.15/0.12/cmember!85!black, 0/0/cextra!85!black}{
    \ccard{2.95+\dx}{\yA-0.15+\dy}{1.0}{1.05}
    \cnet{3.08+\dx}{\yA-0.15+\dy}{0.72}{\col}{stA\j}
}
\node[font=\footnotesize, anchor=north] at (3.55,\yA-0.78)
    {$f_{\vtheta+\sigma\boldsymbol{\epsilon}_i}$};

\draw[flow, -] (baseA-e) -- (plusA.west);
\draw[flow] (epsA.south) -- (plusA.north);
\draw[flow] (plusA.east) -- (2.9,\yA-0.15);

\foreach \dx/\dy/\col/\a/\b/\c/\d/\e [count=\j] in
    {0.30/0.24/csoup!85!black/0.20/0.14/0.50/0.30/0.10,
     0.15/0.12/cmember!85!black/0.22/0.52/0.30/0.18/0.12,
     0/0/cextra!85!black/0.55/0.20/0.34/0.14/0.10}{
    \ccard{\xh-0.08+\dx}{\yA-0.15+\dy}{1.02}{0.84}
    \chist{\xh+\dx}{\yA-0.15+\dy}{\col}{\a}{\b}{\c}{\d}{\e}{hA\j}
}
\draw[flow] (4.35,\yA-0.15) -- (\xh-0.12,\yA-0.15);
\node[font=\footnotesize, anchor=north] at (\xh+0.5,\yA-0.62) {logits $\boldsymbol{\ell}_i$};

\node[note, text=cthicket!80!black] at (6.9,\yA-0.15)
    {$\times N$ perturbed copies\\of all $d_{\vtheta}\!\sim\!10^{9}$ weights};

\cnet{0.45}{\yB-0.15}{0.75}{black!70}{baseB}
\clock{0.83}{\yB+0.55}
\node[font=\footnotesize, anchor=north] at (baseB-s) {$f_{\vtheta}$ (frozen)};

\chist{2.0}{\yB-0.15}{black!70}{0.50}{0.22}{0.42}{0.14}{0.10}{hB0}
\node[font=\footnotesize, anchor=north] at (hB0-s) {$\boldsymbol{\ell}$};
\draw[flow] (baseB-e) -- (hB0-w);

\node[plus] (plusB) at (3.55,\yB-0.15) {$+$};
\node[noise] (epsB) at (3.55,\yB+0.64) {$\sigma\boldsymbol{\epsilon}_i \in \mathbb{R}^{V}$};
\node[font=\scriptsize, text=black!55, anchor=north] at (plusB.south) {into logits};

\draw[flow, -] (hB0-e) -- (plusB.west);
\draw[flow] (epsB.south) -- (plusB.north);

\foreach \dx/\dy/\col/\a/\b/\c/\d/\e [count=\j] in
    {0.30/0.24/csoup!85!black/0.40/0.18/0.52/0.10/0.14,
     0.15/0.12/cmember!85!black/0.42/0.24/0.36/0.18/0.46,
     0/0/cextra!85!black/0.56/0.16/0.40/0.20/0.08}{
    \ccard{\xh-0.08+\dx}{\yB-0.15+\dy}{1.02}{0.84}
    \chist{\xh+\dx}{\yB-0.15+\dy}{\col}{\a}{\b}{\c}{\d}{\e}{hB\j}
}
\draw[flow] (plusB.east) -- (\xh-0.12,\yB-0.15);
\node[font=\footnotesize, anchor=north] at (\xh+0.5,\yB-0.62)
    {$\boldsymbol{\ell}+\sigma\boldsymbol{\epsilon}_i$};

\node[note, text=clogit!80!black] at (6.9,\yB-0.15)
    {one frozen network,\\$\times N$ logit biases,\\size $V\!\sim\!10^{5}$};

\def\yc{-1.425}
\def\gap{0.85}
\node[band, fill=cwrite!7, minimum width=3.3cm, minimum height=5.95cm] 
    (sel) at (10.7,{\yA+1.55}) {};
\node[band, fill=cttmv!8, minimum width=2.9cm, minimum height=5.95cm]
    (vote) at (14.45,{\yA+1.55}) {};

\node[font=\bfseries, anchor=north] at ([yshift=-2mm]sel.north) {Select};
\node[font=\footnotesize, text=black!60, anchor=north] at ([yshift=-6mm]sel.north)
    {\shortstack{Retain top $K$ \\ on validation set}};

\node[draw=black!20, fill=white, rounded corners=2pt, minimum width=2.9cm,
      minimum height=3.35cm, inner sep=0pt] (rankbox) at (12.35,\yc-0.27) {};

\foreach \k/\col/\a/\b/\c/\d/\e/\acc/\keep in
    {1/cextra!85!black/0.40/0.12/0.22/0.08/0.06/0.82/1,
     0/cmember!85!black/0.10/0.38/0.14/0.26/0.08/0.51/0,
    -1/csoup!85!black/0.16/0.08/0.40/0.20/0.10/0.78/1}{
    \pgfmathsetmacro{\yy}{\yc+\k*\gap}
    \ifnum\keep=1
        \node[draw=cwrite!70!black, fill=cwrite!12, rounded corners=2pt,
              minimum width=2.6cm, minimum height=0.6cm, inner sep=0pt]
            at (12.4,\yy) {};
    \fi
    \chist{11.5}{\yy+0.1}{\col}{\a}{\b}{\c}{\d}{\e}{rk}
    \node at (13.35,\yy) {$\acc$};
}
\node[font=\footnotesize\bfseries, text=black!60] at (11.70,\yc-1.6) {Logits};
\node[font=\footnotesize\bfseries, text=black!60, align=center] at (13.10,\yc-1.6)
    {Validation\\accuracy};

\draw[flow] (10.2,\yA) to[out=0, in=180] (sel.west |- 0,\yA);
\draw[flow] (10.2,\yB) to[out=0, in=180] (sel.west |- 0,\yB);

\node[font=\bfseries, anchor=north] at ([yshift=-2mm]vote.north) {Vote};
\node[font=\footnotesize, text=black!60, anchor=north] at ([yshift=-6mm]vote.north)
    {Aggregate on test};

\chist{14.75}{\yc+\gap+0.1}{cextra!85!black}{0.38}{0.10}{0.24}{0.08}{0.06}{vtop}
\chist{14.75}{\yc+0.1}{cmember!35}{0.20}{0.30}{0.14}{0.16}{0.08}{vmid}
\chist{14.75}{\yc-\gap+0.1}{csoup!85!black}{0.34}{0.08}{0.30}{0.16}{0.10}{vbot}

\draw[flow] (13.95,\yc+\gap) -- (14.67,\yc+\gap);
\draw[flow, opacity=0.2] (13.95,\yc) -- (14.67,\yc);
\draw[flow] (13.95,\yc-\gap) -- (14.67,\yc-\gap);

\node[circle, thick, draw=black!35, fill=white, minimum size=6mm, inner sep=1pt]
    (agg) at (16.5,\yc) {\scriptsize\texttt{vote}};
\draw[flow] (15.69,\yc+\gap) -- ++(0.15,0) |- (agg.west);
\draw[flow] (15.69,\yc-\gap) -- ++(0.15,0) |- (agg.west);

\node[inner sep=1pt, rounded corners=2pt, draw=black!40, fill=white,
      minimum width=9mm, minimum height=6mm, font=\bfseries]
    (pred) at (16.5,\yc-1.75) {\footnotesize prediction};
\draw[flow] (agg.south) -- (agg.south |- pred.north);
\end{tikzpicture}%
}

%% file: sections/2_related_work.tex
\section{Related Work}
\label{sec:related}

\textbf{Test-time sampling, selection, and voting.}
Inference-time scaling improves fixed models by generating multiple candidates and aggregating or selecting them. Self-consistency votes over sampled reasoning paths~\citep{wang2023selfconsistency}, while best-of-$N$ selects candidates using a verifier or reward model~\citep{cobbe2021verifiers}; more generally, increasing inference-time compute can substantially improve performance~\citep{snell2024testtime}. Perturb-and-MAP provides a classical connection between perturbation and sampling, where random logit perturbations followed by MAP inference induce samples from a target distribution~\citep{papandreou2011perturb,maddison2017concrete}. Neural Thickets~\citep{gan2026thickets} combines these ingredients: perturbations of the weights generate candidate models, validation performance selects them, and voting aggregates their predictions.

\textbf{Logit-space control.}
Many methods alter the behavior of a frozen model directly through its logits, including classifier-free guidance~\citep{ho2022cfg}, contrastive decoding~\citep{li2023contrastive}, post-hoc calibration~\citep{guo2017calibration}, and watermarking, which biases a pseudo-random subset of token logits to embed a detectable signal~\citep{kirchenbauer2023watermark}. These methods design or fit a particular logit transformation for a desired behavior. In contrast, logit PSV uses task-agnostic random biases and relies on selection and voting to exploit the resulting output diversity.

\textbf{Perturbations in weight space.}
Previous work has revealed substantial structure in the local geometry of neural-network parameters, including mode connectivity~\citep{garipov2018mode}, model averaging~\citep{izmailov2018swa,wortsman2022soups}, and low-dimensional adaptation~\citep{li2018intrinsic,hu2022lora}. Random weight perturbations are also central to evolution strategies and zeroth-order optimization~\citep{salimans2017es}. Our focus is different: rather than optimizing or merging perturbations, we ask which aspects of weight-space perturbation are actually necessary for the gains of PSV.
Additional related work is discussed in Appendix~\ref{app:extended_related}.

%% file: sections/3_preliminaries.tex
\section{The Perturb-Select-Vote Framework}
\label{sec:prelim}

\begin{figure}[t]
\centering
\input{figures/fig_perlayer}
\caption{\textbf{Effect of perturbing individual decoder blocks.}
Countdown accuracy with OLMo-3-7B-Instruct ($K=50$), perturbing one block at a time while keeping all other weights fixed.}
\label{fig:perlayer}
\end{figure}

We use \emph{perturb-select-vote} (PSV) to denote a general test-time adaptation template inspired by Neural Thickets, but not restricted to full-weight perturbations. The procedure is defined by three choices: a perturbation space, a selection set, and an aggregation rule. Neural Thickets~\citep{gan2026thickets} correspond to the full-weight instance of PSV, where the perturbation space is the full parameter vector of the model, see top path of Figure~\ref{fig:pipeline}.

\paragraph{Perturb-select-vote.}
Let $f\left(\cdot;\vtheta\right)$ be a pretrained predictor. PSV samples $N$ perturbed predictors
\begin{align}
    f_i
    =
    \mathcal{P}
    \left(
        f\left(\cdot;\vtheta\right);
        \boldsymbol{\epsilon}_i
    \right),
    \qquad
    \boldsymbol{\epsilon}_i
    \sim
    \mathcal{D}_{\sigma},
    \qquad
    i = 1,\dots,N,
    \label{eq:psv_generic}
\end{align}
where $\mathcal{P}$ is a perturbation operator and $\mathcal{D}_{\sigma}$ is a noise distribution with scale $\sigma$. Each member is evaluated on a task-specific selection set, yielding a reward $r_i$. The top $K$ members are retained:
\begin{align}
    \mathcal{S}
    =
    \operatorname{TopK}
    \left(
        \left\{r_i\right\}_{i=1}^{N}
    \right)~,
    \label{eq:psv_select}
\end{align}
and the final prediction is obtained by aggregating the predictions of members $i \in \mathcal{S}$, typically by majority vote. The case $K=1$ corresponds to selecting the single best perturbation.

\begin{figure}[t]
\centering
\input{figures/fig_block}
\caption{\textbf{Single-component perturbations within a decoder block.}
Fraction of the full-weight PSV gain recovered by perturbing one component of the first decoder block of OLMo-3-7B-Instruct on Countdown with $K=50$. All components recover at least $85\%$ of the full-weight gain (base $66.6\%$, full-weight PSV $90.8\%$).}
\label{fig:block}
\end{figure}

\paragraph{Neural Thickets as full-weight PSV.}
RandOpt, the method introduced by~\citet{gan2026thickets}, instantiates PSV by perturbing every model parameter. For a model with parameters $\vtheta \in \mathbb{R}^{d_{\vtheta}}$ where $d_{\vtheta}$ is the number of parameters in $f$, each member is
\begin{align}
    f_i
    =
    f
    \left(
        \,\cdot\,;
        \vtheta
        +
        \sigma \boldsymbol{\epsilon}_i
    \right),
    \qquad
    \boldsymbol{\epsilon}_i
    \sim
    \mathcal{N}
    \left(
        \mathbf{0},
        \mathbf{I}_{d_{\vtheta}}
    \right).
    \label{eq:randopt}
\end{align}
Each perturbed model is scored on a selection set of $200$ target-task examples, and the top $K$ models are ensembled by majority vote. Following~\citet{gan2026thickets}, we consider perturbation budgets from $N=1000$ to $N=5000$ and ensemble sizes $K \in \left\{1,2,5,10,20,50\right\}$.

\paragraph{Empirical effect.}
The original Neural Thickets study evaluates RandOpt on OLMo-3-7B-Instruct~\citep{olmo2024}, {OLMo-3-7B,} Qwen2.5-0.5B-Instruct, {Qwen2.5-1.5B-Instruct, Qwen2.5-3B-Instruct~\citep{qwen2025}, and Llama-3.1-8B-Instruct}, across seven tasks: countdown, GSM8K, MATH-500, OlympiadBench, MBPP, ROCStories, and USPTO. {For OLMo-3-7B-Instruct at $K=50$ and $N=5000$, the reported gains over greedy decoding are $+20.2$ points on countdown, $+15.8$ on USPTO, $+13.1$ on MATH-500, $+9.2$ on MBPP, $+6.7$ on OlympiadBench, and $+6.6$ on GSM8K. The gain is smaller on ROCStories $\left(+0.5\right)$, which the base model already handles well. Running the authors' released code in the same setting, we obtain $+25.2$, $+14.0$, $+19.0$, $+10.4$, $+11.0$, $+2.1$, and $+0.7$; the final accuracies are within $3.7$ points of the reported ones on six tasks and $6.8$ points higher on countdown (Appendix~\ref{app:upstream}). The other language experiments in this paper use a variant of that setting: generations of up to $2048$ tokens on MATH-500, OlympiadBench, and MBPP and $64$ on ROCStories (instead of $1024$), and different OlympiadBench and ROCStories splits. This raises the base accuracy on MATH-500 and OlympiadBench, and every comparison is made within one setting.} Thus, our work does not dispute the empirical phenomenon: full-weight PSV can produce large and practically meaningful gains.

\paragraph{Weight-space interpretation.}
The central interpretation of~\citet{gan2026thickets} is that these gains reveal local task experts near the pretrained weights. This interpretation rests on three main claims. First, a \emph{density} claim: as model size increases, a larger fraction of nearby perturbations improves task performance. Second, a \emph{specialist} claim: perturbations that improve one task tend to hurt others, suggesting task-specific specialization. Third, a \emph{competitiveness} claim: RandOpt with $K=50$ mostly matches or outperforms task-adaptation methods, including reinforcement-learning methods such as PPO~\citep{schulman2017proximal} and GRPO~\citep{shao2024deepseekmath}, while a test-time majority-vote baseline is reported as weaker. Together, these claims frame the gains as a property of the local weight neighborhood. In the following sections, we keep the PSV procedure fixed and vary only the perturbation space, asking whether the observed gains require perturbing the full weight vector.

%% file: figures/fig_perlayer.tex
\begin{tikzpicture}
\begin{axis}[
paperaxis,
width=0.8\linewidth,
height=4.3cm,
xlabel={index of perturbed decoder block},
ylabel={countdown acc.(\%)},
  xmin=-1,
  xmax=32,
  ymin=63,
  ymax=95,
  xtick={0,5,10,15,20,25,30},
  legend style={
      at={(1.0,0.3)},
      anchor=east,
      font=\footnotesize
    }]

\addplot[clogit, mark=*, mark size=1.1pt, line width=0.6pt] coordinates {
  (0,85.2)(1,76.0)(2,76.0)(3,80.0)(4,79.5)(5,76.8)(6,77.5)(7,78.2)(8,82.7)(9,81.0)(10,82.0)(11,83.0)(12,83.8)(13,81.9)(14,82.1)(15,82.3)(16,83.1)(17,82.8)(18,78.3)(19,85.9)(20,90.0)(21,88.6)(22,89.3)(23,87.8)(24,89.2)(25,89.2)(26,89.8)(27,89.5)(28,89.5)(29,89.8)(30,89.7)(31,91.8)};
\addlegendentry{single block perturbation}
\addplot[cthicket, densely dotted, line width=0.8pt] coordinates {(-1,90.8)(32,90.8)} node[pos=0.03, below, anchor=north west, font=\footnotesize] {full weight perturbation};

\addplot[cbase, dashed, line width=0.8pt]
coordinates {(-1,66.6)(32,66.6)}
node[
  pos=0.03,
  above=-1mm,
  anchor=south west,
  font=\footnotesize
] {no perturbation};

\end{axis}
\end{tikzpicture}

%% file: figures/fig_block.tex
\resizebox{\linewidth}{!}{%
\begin{tikzpicture}[
  font=\small, >=Stealth, line cap=round, line join=round,
  comp/.style={draw, rounded corners=2.5pt, minimum width=1.6cm, minimum height=0.8cm, align=center, line width=0.6pt, inner sep=1pt},
  write/.style={comp, fill=cwrite!30, draw=cwrite!70!black},
  qkn/.style={comp, fill=cttmv!30, draw=cttmv!75!black},
  cmp/.style={comp, fill=cthicket!25, draw=cthicket!70!black},
  nlbox/.style={draw, dashed, rounded corners=4pt, fill=cbase!12, draw=black!40, minimum width=1.1cm, minimum height=0.75cm, inner sep=0pt, font=\footnotesize\itshape, text=black!60, align=center},
  add/.style={draw, circle, fill=white, inner sep=1.4pt, font=\small, line width=0.7pt},
  fl/.style={->, black!45, line width=0.5pt},
  wr/.style={->, cwrite!80!black, line width=1.5pt},
  card/.style={draw=black!18, rounded corners=5pt, line width=0.6pt},
]
\def\W#1{\textcolor{cwrite!45!black}{\bfseries #1}}
\def\Q#1{\textcolor{cttmv!55!black}{\bfseries #1}}
\def\C#1{\textcolor{cthicket!55!black}{\bfseries #1}}
\def\swatch#1{\tikz[baseline=-0.6ex]\draw[#1, rounded corners=1.5pt] (0,-0.13) rectangle (0.3,0.17);}

\draw[line width=2.6pt, cthicket!80, loosely dotted] (-0.4,0) coordinate (input) -- (0.3,0) coordinate (startofstream);
\draw[line width=2.6pt, cthicket!80, ->] (startofstream) -- (17.3,0) coordinate (output);
\node[cthicket!55!black, anchor=south west, inner sep=1pt] at ([yshift=1mm]input) {input};
\node[cthicket!55!black, anchor=north east, inner sep=1pt] at ([yshift=-3.8mm]output) {logits $\boldsymbol{\ell}\approx\mathbf{W}_U\mathrm{LN}(\mathbf{h})$};
\node[cthicket!55!black, anchor=north west, inner sep=1pt] at ([yshift=-1.5mm]input) {residual stream $\mathbf{h}$};
\node[add] (s1) at (8.20,0) {$+$};
\node[add] (s2) at (16.00,0) {$+$};

\node[cmp] (qp) at (0.80,3.10) {\texttt{q\_proj}\\[1pt]\C{88.8\%}};
\node[cmp] (kp) at (0.80,2.20) {\texttt{k\_proj}\\[1pt]\C{88.0\%}};
\node[cmp] (vp) at (0.80,1.30) {\texttt{v\_proj}\\[1pt]\C{96.3\%}};
\node[qkn] (qn) at (2.70,3.10) {\texttt{q\_norm}\\[1pt]\Q{85.1\%}};
\node[qkn] (kn) at (2.70,2.20) {\texttt{k\_norm}\\[1pt]\Q{87.6\%}};
\node[add, scale=0.75, inner sep=0.25pt] (product) at (3.85,2.65) {$\times$};
\node[nlbox] (sm) at (4.80,2.65) {softmax};
\node[write] (oo) at (6.40,2.65) {\texttt{o\_proj}\\[1pt]\W{98.8\%}};
\node[write] (la) at (8.20,2.65) {LN$_{\text{attn}}$\\[1pt]\W{103.3\%}};

\node[cmp] (gp) at (10.75,2.65) {\texttt{gate\_proj}\\[1pt]\C{96.7\%}};
\node[cmp] (up) at (10.75,1.75) {\texttt{up\_proj}\\[1pt]\C{99.2\%}};
\node[nlbox] (sw) at (12.60,2.20) {SwiGLU};
\node[write] (dp) at (14.20,2.20) {\texttt{down\_proj}\\[1pt]\W{100.8\%}};
\node[write] (lf) at (16.00,2.20) {LN$_{\text{ff}}$\\[1pt]\W{100.8\%}};

\begin{scope}[on background layer]
  \node[card, fit=(qp)(vp)(la), inner sep=5pt] (abox) {};
  \node[card, fit=(gp)(up)(lf), inner sep=5pt] (mbox) {};
\end{scope}
\node[black!50, anchor=south west, font=\small\itshape, inner sep=1pt] at ([xshift=4pt]abox.north west) {self-attention};
\node[black!50, anchor=south west, font=\small\itshape, inner sep=1pt] at ([xshift=4pt]mbox.north west) {feed-forward};

\draw[fl] (qp) -- (qn);
\draw[fl] (kp) -- (kn);
\draw[fl] (qn.east) -| (product);
\draw[fl] (kn.east) -| (product);
\draw[fl] (product) -- (sm.west);
\draw[fl] (vp.east) -| (sm.south);
\draw[fl] (sm.east) -- (oo.west);
\draw[fl] (oo.east) -- (la.west);
\draw[fl] (gp.east) -- ++ (0.3,0) |- (sw) coordinate[midway] (tmp) ;
\draw[fl,-] (up.east) -- ++ (0.3,0) |- (tmp);
\draw[fl] (sw.east) -- (dp.west);
\draw[fl] (dp.east) -- (lf.west);

\draw[wr, ->] (vp |- startofstream) -- (vp);
\draw[wr, ->] (up |- startofstream) -- (up);
\draw[wr, ->] (la.south)  -- (s1) ;
\draw[wr, ->] (lf.south) -- (s2);
\node[black!50, font=\footnotesize, anchor=west] at (0.88,0.42) {read};
\node[cwrite!45!black, font=\footnotesize, anchor=west] at (8.28,0.42) {write};

\node[anchor=north, font=\footnotesize, inner sep=0pt] at (8.45,-1.05) {%
  \swatch{fill=cwrite!30, draw=cwrite!70!black}\ writes into the stream\ (\W{$\approx{+}24$})\qquad
  \swatch{fill=cttmv!30, draw=cttmv!75!black}\ QK-norm: attention temperature\ (\Q{$\approx{+}21$})\qquad
  \swatch{fill=cthicket!25, draw=cthicket!70!black}\ compute matrices\ (\C{${+}21$ to ${+}24$})};
\end{tikzpicture}}

%% file: sections/6_minimal_method.tex
\section{Weight Perturbations Are Not Necessary}
\label{sec:minimal}

We now investigate the mechanism underlying PSV. We progressively simplify the perturbation mechanism, first localizing it within the network and then moving it entirely to the output logits. This output-space view connects PSV to standard sampling, selection, and voting methods, allows us to isolate the source of its gains, and finally explains why internal perturbations behave similarly through the logit perturbations they induce.

\subsection{Local Weight Perturbations Suffice}
\label{sec:local}

Full-weight PSV, as used in Neural Thickets, perturbs every parameter of the model. However, such exhaustive perturbations are not necessary. On OLMo-3-7B-Instruct for Countdown with $K=50$, the greedy base model reaches $66.6\%$ and full-weight PSV reaches $90.8\%$. As shown in Figure~\ref{fig:perlayer}, perturbing only a single decoder block already recovers much of this gain. Every block substantially improves over the base model, with later blocks being the most effective. Thus, the gain does not require a coordinated perturbation of the full network.

The same holds within a transformer block. We perturb a single component of the first decoder block while keeping all remaining parameters frozen. Every component recovers most of the full-weight gain (Figure~\ref{fig:block}). Components that write directly into the residual stream recover the full gain, while the value and feed-forward input projections recover $96$--$99\%$. The query/key path, which acts indirectly through the attention softmax, recovers $85$--$89\%$. Qwen2.5 exhibits the same pattern from 0.5B to 3B (Appendix~\ref{app:cross}). Thus, most of the PSV gain can arise from highly localized perturbations.

\subsection{Perturbing Only the Logits}
\label{sec:logit}

\begin{figure}[t]
\centering
\input{figures/fig_logitbias}
\caption{\textbf{Logit PSV matches full-weight PSV across tasks and model scales.}
\textit{(a)} On OLMo-3-7B-Instruct, logit PSV matches or exceeds full-weight PSV on all evaluated tasks.
\textit{(b)} Across model scales and tasks, logit PSV closely tracks full-weight PSV.}
 \label{fig:logitbias}
 \end{figure}

The preceding results raise a more direct question: are weight perturbations needed at all? We test this by moving the perturbation entirely to the output space. The model remains fully frozen, and only its output logits are perturbed during decoding. This yields a simpler PSV variant and connects PSV directly to standard sampling and ensembling methods.

Let $f\left(\cdot;\vtheta\right)$ be the frozen pretrained model, and let $\boldsymbol{\ell} \in \mathbb{R}^{V}$ denote its logits over the vocabulary $V$ for a given token. Each PSV member $i$ draws a vocabulary-sized Gaussian bias vector and decodes greedily under the perturbed logits, as shown in the bottom path of Figure~\ref{fig:pipeline}:
\begin{align}
    \boldsymbol{\ell}_i
    =
    \boldsymbol{\ell}
    +
    \sigma \boldsymbol{\epsilon}_i,
    \qquad
    \boldsymbol{\epsilon}_i
    \sim
    \mathcal{N}
    \left(
        \mathbf{0},
        \mathbf{I}_{V}
    \right),
    \label{eq:logitbias}
\end{align}
where $V$ is the vocabulary size and $\sigma>0$ controls the noise scale. The bias $\boldsymbol{\epsilon}_i$ is fixed for member $i$ {across all decoding steps and inputs}. As in full-weight PSV, we sample $N$ members, score each one on the selection set, retain the top $K$, and return the majority vote at test time. This variant, which we call \emph{logit PSV}, modifies neither the weights nor the hidden states of the model.%

Despite this restriction, logit PSV matches or exceeds full-weight PSV on most reasoning tasks, as shown in Figure~\ref{fig:logitbias}. On OLMo-3-7B-Instruct, it raises MATH-500 from $63.3\%$ to {$81.3\%$}, compared with $74.7\%$ for full-weight PSV at the same $N$ and $K$. It also improves GSM8K from $87.5\%$ to {$91.9\%$}. {On USPTO it reaches $42.7\%$ against $39.4\%$, but by predicting the majority reaction class more often: its balanced accuracy is lower ($26.7\%$ against $29.7\%$).} Across the 0.5B, 1.5B, 3B, and 7B models, logit PSV closely tracks the full-weight method{: it is more accurate on $13$ of the $15$ reasoning cells and within $0.6$ points on the other two, while on MBPP, ROCStories, and USPTO it is usually below it} (Table~\ref{tab:crossscale}). {Most of these differences are within test-set noise. At a matched budget of $N{=}5000$ under the original protocol, the two methods are statistically indistinguishable on $22$ of $28$ model--task pairs; logit PSV is significantly better on three reasoning pairs and significantly worse on three others, and never worse on a reasoning task (Appendix~\ref{app:upstream}).} Since this intervention performs no computation inside the network, these results show that the gains of full-weight PSV can be reproduced by injecting diversity directly in output space.

\subsection{Connection to Sampling, Selection, and Voting}
\label{sec:theory}

\begin{figure}[t]
\centering
\input{figures/fig_ttmv}
\caption{\textbf{Temperature-tuned majority voting versus full-weight PSV.}
(\textit{a}) On MATH-500 with a strong model, tuned TT-MV matches or exceeds full-weight PSV.
(\textit{b}) On USPTO with a weaker model, TT-MV remains far below PSV, showing the value of validation-based selection; both remain below the majority-class baseline.
(\textit{c}) Across tasks and model scales, TT-MV at $T{=}0.7$ closely tracks full-weight PSV, with larger gaps for weaker models and harder tasks.}
 \label{fig:ttmv}
\end{figure}

Logit PSV admits a simple interpretation in terms of three familiar inference-time mechanisms. We make these connections precise below and contrast them with temperature sampling and best-of-$N$.

\paragraph{A fixed logit bias induces persistent token preferences.}
Let $x$ denote the input, $y_{<t}$ the sequence generated up to step $t$, and
$p\left(v \mid x,y_{<t}\right)$ the next-token distribution of the frozen model.
For ensemble member $i$, we sample once a bias
$\boldsymbol{\epsilon}_i \in \mathbb{R}^V$ and add it to the logits at every decoding step.
The resulting next-token distribution is
\begin{align}
    p_i\left(v \mid x,y_{<t}\right)
    &\propto
    {
        p\left(v \mid x,y_{<t}\right)
        \exp\left(\sigma \epsilon_{i,v}\right)
    }~.
    \label{eq:recalib}
\end{align}
Thus, $\epsilon_{i,v}>0$ systematically increases member $i$'s relative preference for token $v$, while $\epsilon_{i,v}<0$ decreases it. Sampling a bias and then greedily decoding is therefore a form of perturb-and-MAP sampling~\citep{papandreou2011perturb}: randomness is injected into the scores before taking an argmax. It differs, however, from standard temperature sampling, which can be viewed as drawing a new random perturbation at each autoregressive step. In logit PSV, the same bias $\boldsymbol{\epsilon}_i$ is reused throughout the entire decoding trajectory and across all inputs evaluated by member $i$. Setting $\sigma=0$ recovers ordinary greedy decoding.

\paragraph{Voting corresponds to self-consistency over random perturbations.}
For a fixed input $x$, each sampled bias $\boldsymbol{\epsilon}$ defines one greedily decoded answer $a_{\boldsymbol{\epsilon}}(x)$. Since the decoding itself is deterministic once $\boldsymbol{\epsilon}$ is fixed, sampling biases induces a distribution over complete answers:
\begin{align}
    q_{\sigma}\left(a \mid x\right)
    &=
    \mathrm{Pr}_{\boldsymbol{\epsilon}}
    \left[
        a_{\boldsymbol{\epsilon}}(x)=a
    \right]~.
    \label{eq:answer_distribution}
\end{align}
In other words, $q_{\sigma}\left(a \mid x\right)$ is the probability that a random logit bias leads to answer $a$. Before selection, different PSV members correspond to independent samples from this distribution. Majority voting therefore plays the same role as self-consistency~\citep{wang2023selfconsistency}: it returns the answer produced most often across independently perturbed members.

\paragraph{Selection retains perturbations that transfer across inputs.}
The key difference between logit PSV and ordinary output sampling is that each sampled bias defines a persistent ensemble member. Its validation reward therefore measures how well the same token-preference shift performs across many examples from the target task. PSV ranks the sampled biases by this reward, retains the top $K$, and reuses those same biases on unseen test inputs.

This differs from best-of-$N$ decoding~\citep{cobbe2021verifiers}, where multiple candidate answers are generated independently for each test input and a verifier or reward model selects among them at test time. PSV instead performs selection once on a validation set and requires no test-time verifier. It also differs from temperature sampling, whose randomness is resampled independently at each decoding step and for each input; individual random draws therefore do not define reusable members whose validation performance can be transferred to future examples (see Proposition~\ref{prop:persist}). Temperature-tuned majority voting (TT-MV), the test-time baseline of~\citet{gan2026thickets} built on self-consistency~\citep{wang2023selfconsistency}, shares the output diversity and voting stages of logit PSV, but omits this validation-based selection of persistent perturbations.

\subsection{Disentangling the Source of PSV Gains}
\label{sec:ttmv}

We next ask which parts of PSV are necessary. To isolate the effect of output diversity and voting without validation-based selection, we compare against temperature-tuned majority voting (TT-MV). This baseline majority-votes over samples of the base model at temperature $T$. {We fix $T{=}0.7$ for every task and model instead of choosing it on the test set; choosing it on the test set would raise TT-MV by at most $4.0$ points on OLMo-3-7B-Instruct, but by up to $16.9$ points on weaker models.}

\paragraph{Test-time sampling and majority voting.}
Figure~\ref{fig:ttmv} and Table~\ref{tab:master} show that, on strong base models, TT-MV explains much of the gain. On MATH-500 with OLMo-3-7B-Instruct, TT-MV {reaches} $77.7\%$, above full-weight PSV at the matched budget $\left(74.7\%\right)$. {It exceeds the $N=1000$ full-weight PSV on three of seven tasks (Countdown, MATH-500, and OlympiadBench), is within $0.4$ points on GSM8K and ROCStories, and trails it by $2.1$ points on MBPP and $3.1$ on USPTO.} It also exceeds the $N=5000$ full-weight PSV on Countdown {and OlympiadBench}, despite using a smaller search budget.

\begin{table}[t]
\begin{minipage}{0.32\linewidth}
\centering
\caption{\textbf{Comparison on OLMo-3-7B-Instruct.}
Accuracy at $K=50$ for the base model, model soup, full-weight PSV ($N=5000$), TT-MV ($T=0.7$), and logit PSV ($N=1000$); the model soup also uses $N=1000$. Despite a $5\times$ smaller search budget, logit PSV matches or exceeds full-weight PSV on five of seven tasks; TT-MV does so on two.
}
\label{tab:master}
\end{minipage}
\hfill
\begin{minipage}{0.63\linewidth}
\input{tables/tab_master}

\end{minipage}
\end{table}

\paragraph{Selection matters when unfiltered samples are unreliable.}
TT-MV removes the validation-selection step, so its failures identify regimes where diversity and voting alone are insufficient. On weaker base models, sampling can produce diverse outputs without assigning enough probability mass to correct ones. Figure~\ref{fig:ttmv} shows such a case on Qwen2.5-1.5B-Instruct for USPTO: TT-MV remains at or below $6.9\%$ for all temperatures, while full-weight PSV reaches $24.9\%$. This gain largely reflects improved label consistency rather than better task performance: PSV predicts the majority reaction class on $76\%$ of test examples and still underperforms the $31.9\%$ majority-class baseline. A cleaner example is Countdown with the same model, where TT-MV peaks at $34.3\%$ ($T{=}0.12$), below full-weight PSV at $37.4\%$. This gap does not require a weight-space expert explanation. It shows that validation-based selection can be important when the base sampler is too unreliable for majority voting alone.
Overall, PSV gains arise from candidate diversity, validation-based selection, and voting. Strong models may benefit from sampling and voting alone, whereas weaker models also benefit from selection.

\subsection{Internal Perturbations Induce Logit Perturbations}
\label{sec:linearized_psv}

We provide a linearized view of PSV that connects full-weight perturbations to output-space perturbations. In this view, a weight perturbation matters through its induced first-order change in the logits, and logit PSV directly injects a simpler version of the same output diversity.

\paragraph{Weight perturbations induce first-order logit perturbations.}
Let $\boldsymbol{\ell}_{\boldsymbol{\theta}}\left(x\right)$ denote the logits of a pretrained model at input $x$. For a perturbed model $\boldsymbol{\theta}+\sigma\boldsymbol{\epsilon}$, a first-order expansion gives
\begin{align}
\boldsymbol{\ell}_{\boldsymbol{\theta}+\sigma\boldsymbol{\epsilon}}\left(x\right)
    \approx
    \boldsymbol{\ell}_{\boldsymbol{\theta}}\left(x\right)
    +
    \sigma
    \mathbf{J}\left(x\right)
    \boldsymbol{\epsilon},
    \qquad
    \mathbf{J}\left(x\right)
    =
    \frac{\partial \boldsymbol{\ell}_{\boldsymbol{\theta}}\left(x\right)}
    {\partial \boldsymbol{\theta}}~.
\end{align}
Thus, regardless of where the perturbation is applied in the network, its leading-order effect is a structured perturbation of the logits. {This explains why perturbing even a small subset of the weights can recover most of the effect of perturbing the full network (Figure~\ref{fig:block}): any internal perturbation is propagated through the residual stream to the logits, while its location mainly determines the magnitude and structure of the resulting logit perturbation. In particular, the attention softmax attenuates perturbations along the query/key path without eliminating them (Appendix~\ref{app:residual}).} Logit PSV can therefore be viewed as a cheaper output-space surrogate: it injects diversity directly at the logits, rather than producing it indirectly through a forward pass with perturbed weights.

We empirically validate this approximation on OLMo-3-7B-Instruct for Countdown. At the perturbation scale used in our experiments, the linearized logits reproduce $89.7\%$ of the argmax-token changes induced by the exact perturbed model; agreement is higher at smaller scales and degrades as $\sigma$ increases, as expected from a first-order approximation (Table~\ref{tab:linear_approx}). This supports the interpretation that PSV selection and voting act largely on perturbation-induced logit shifts, rather than on a qualitatively different internal computation.

\begin{table}[t]
    \centering
    \begin{minipage}[t]{0.5\linewidth}
    \vspace{0pt}
    \caption{\textbf{Weight perturbations are well approximated by first-order logit shifts.}
    On OLMo-3-7B-Instruct for Countdown, we compare exact weight perturbations to their first-order logit approximation. We report the fraction of argmax-token flips induced by the exact perturbation that are reproduced by the linearized logits. The perturbation scale used in our experiments is shown in bold.}
    \label{tab:linear_approx}
    \end{minipage}
    \hfill
    \begin{minipage}[t]{0.45\linewidth}
    \vspace{5pt}
    \input{tables/linear}
    \end{minipage}
\end{table}

%% file: figures/fig_logitbias.tex
\begin{minipage}[t]{0.50\linewidth}
\centering

\begin{tikzpicture}
\begin{axis}[paperaxis, width=\linewidth, height=4.7cm,
  ybar, bar width=3.4pt, enlarge x limits=0.06,
  symbolic x coords={countdown,GSM8K,MATH-500,Olympiad,MBPP,ROCStories,USPTO},
  xtick=data, x tick label style={rotate=35, anchor=north east, font=\scriptsize, inner sep=1pt},
  ymin=0, ymax=100, ylabel={accuracy @\,$K{=}50$ (\%)},
  legend columns=3, legend style={at={(0.5,1.02)}, anchor=south}]
\addplot[fill=cbase, draw=cbase!60!black]      coordinates {(countdown,66.6)(GSM8K,87.49)(MATH-500,63.33)(Olympiad,37.97)(MBPP,64.27)(ROCStories,65.2)(USPTO,28.57)}; \addlegendentry{base}
\addplot[fill=cthicket, draw=cthicket!60!black] coordinates {(countdown,90.75)(GSM8K,89.76)(MATH-500,74.67)(Olympiad,46.2)(MBPP,76.44)(ROCStories,65.57)(USPTO,39.36)}; \addlegendentry{full-weight PSV}
\addplot[fill=clogit, draw=clogit!60!black]     coordinates {(countdown,90.70)(GSM8K,91.89)(MATH-500,81.33)(Olympiad,52.53)(MBPP,73.17)(ROCStories,65.76)(USPTO,42.66)}; \addlegendentry{logit PSV}
\end{axis}
\end{tikzpicture}
\footnotesize \textbf{(a) Logit PSV matches full thicket performance}
\end{minipage}
\hfill
\begin{minipage}[t]{0.45\linewidth}
\centering
\tikzset{
  model7B/.style   ={cthicket, mark options={fill=cthicket, draw=cthicket!55!black, line width=0.4pt}},
  model3B/.style   ={cextra, mark options={fill=cextra, draw=cextra!55!black, line width=0.4pt}},
  model15B/.style  ={cttmv, mark options={fill=cttmv, draw=cttmv!55!black, line width=0.4pt}},
  model05B/.style  ={csoup, mark options={fill=csoup, draw=csoup!55!black, line width=0.4pt}},
  taskCD/.style    ={mark=*},
  taskGSM/.style   ={mark=diamond*},
  taskMATH/.style  ={mark=square*},
  taskOLY/.style   ={mark=triangle*},
  taskMBPP/.style  ={mark=pentagon*},
  taskROC/.style   ={mark=star},
  taskUSPTO/.style ={mark=x},
}%
\begin{tikzpicture}
\begin{axis}[
  paperaxis,
  width=\linewidth,
  height=4.7cm,
  xlabel={full-weight thicket (\%)},
  ylabel={logit PSV (\%)},
  xmin=10, xmax=95,
  ymin=10, ymax=95,
  xtick={20,50,80},
  ytick={20,50,80},
  clip=false,
]

\addplot[
  gray,
  dashed,
  line width=0.5pt,
  domain=10:95,
  forget plot
] {x};

\taskpoint{90.75}{90.70}{model7B}{taskCD}
\taskpoint{89.76}{91.89}{model7B}{taskGSM}
\taskpoint{74.67}{81.33}{model7B}{taskMATH}
\taskpoint{46.20}{52.53}{model7B}{taskOLY}
\taskpoint{76.44}{73.17}{model7B}{taskMBPP}
\taskpoint{65.57}{65.76}{model7B}{taskROC}
\taskpoint{39.36}{42.66}{model7B}{taskUSPTO}

\taskpoint{55.15}{56.95}{model3B}{taskCD}
\taskpoint{87.70}{88.70}{model3B}{taskGSM}
\taskpoint{72.30}{71.67}{model3B}{taskMATH}
\taskpoint{38.40}{40.51}{model3B}{taskOLY}
\taskpoint{64.00}{63.09}{model3B}{taskMBPP}
\taskpoint{56.10}{55.28}{model3B}{taskROC}
\taskpoint{19.80}{15.58}{model3B}{taskUSPTO}

\taskpoint{37.40}{37.50}{model15B}{taskCD}
\taskpoint{76.70}{81.95}{model15B}{taskGSM}
\taskpoint{57.70}{60.00}{model15B}{taskMATH}
\taskpoint{28.50}{28.69}{model15B}{taskOLY}
\taskpoint{52.50}{49.08}{model15B}{taskMBPP}
\taskpoint{49.00}{48.22}{model15B}{taskROC}
\taskpoint{24.90}{31.27}{model15B}{taskUSPTO}

\taskpoint{56.00}{56.56}{model05B}{taskGSM}
\taskpoint{39.00}{40.67}{model05B}{taskMATH}
\taskpoint{14.30}{16.66}{model05B}{taskOLY}
\taskpoint{28.80}{26.44}{model05B}{taskMBPP}
\taskpoint{31.90}{31.87}{model05B}{taskROC}
\node[
  anchor=south east,
  font=\tiny,
  inner sep=1pt,
  fill=white,
] at (axis description cs:0.98,0.04) {%
  \begin{tabular}{@{}c@{\hspace{2pt}}l@{}}
    \legendmark[model7B]{taskCD}   & 7B   \\
    \legendmark[model3B]{taskCD}   & 3B   \\
    \legendmark[model15B]{taskCD}  & 1.5B \\
    \legendmark[model05B]{taskCD}  & 0.5B
  \end{tabular}%
};

\node[
  anchor=north west,
  font=\tiny,
  inner sep=0pt,
  align=center,
] at (axis description cs:0.04,0.98) {%
  \begin{tabular}{@{}c@{\hspace{1pt}}l@{\hspace{5pt}}
                     c@{\hspace{1pt}}l@{\hspace{5pt}}
                  @{}}
    \legendmark[black]{taskCD}   & countdown &
    \legendmark[black]{taskGSM}  & GSM8K \\
    \legendmark[black]{taskMATH} & MATH-500 &
    \legendmark[black]{taskOLY}  & Olympiad \\
    \legendmark[black]{taskMBPP} & MBPP &
    \legendmark[black]{taskROC}  & ROCStories \\
    \legendmark[black]{taskUSPTO} & USPTO
  \end{tabular}%
};

\end{axis}
\end{tikzpicture}
\footnotesize \textbf{(b) Logit PSV vs.\ full-thicket across scales}\\[2pt]
\end{minipage}

%% file: figures/fig_ttmv.tex
\centering
\begin{minipage}[t]{0.32\linewidth}
\centering
\begin{tikzpicture}
\begin{axis}[paperaxis, width=\linewidth, height=4.4cm,
  xmode=log, xlabel={temperature $T$}, ylabel={MATH-500 TT-MV (\%)},
  xmin=0.0008, xmax=1.3, ymin=60, ymax=82,
  legend style={at={(1.0,0.15)}, anchor=south east, font=\tiny}]
\addplot[cttmv, mark=*, mark size=1.2pt] coordinates {(0.001,68.67)(0.01,68.67)(0.3,75.33)(0.5,76.67)(0.7,77.67)(1.0,77.00)}; \addlegendentry{TT-MV}
\addplot[cthicket, dashed, line width=0.8pt] coordinates {(0.0008,74.67)(1.3,74.67)}; \addlegendentry{thicket}
\addplot[cbase, densely dotted, line width=0.8pt] coordinates {(0.0008,63.33)(1.3,63.33)}; \addlegendentry{base}
\end{axis}
\end{tikzpicture}
\text{\footnotesize \shortstack{(a) Strong-model \\ temperature sweep}}
\end{minipage}
\hfill
\begin{minipage}[t]{0.32\linewidth}
\centering
\begin{tikzpicture}
\begin{axis}[paperaxis, width=\linewidth, height=4.4cm,
  xmode=log, xlabel={temperature $T$}, ylabel={Qwen-1.5B USPTO (\%)},
  xmin=0.0008, xmax=1.3, ymin=0, ymax=36,
  legend style={at={(1.0,0.22)}, anchor=south east, font=\tiny}]
\addplot[cttmv, mark=*, mark size=1.0pt] coordinates {(0.001,6.8)(0.005,6.8)(0.01,6.8)(0.02,6.8)(0.05,6.9)(0.08,6.8)(0.1,6.8)(0.15,6.5)(0.3,4.7)(0.5,3.0)(0.7,2.3)(1.0,2.8)}; \addlegendentry{TT-MV}
\addplot[cthicket, dashed, line width=0.8pt] coordinates {(0.0008,24.9)(1.3,24.9)}; \addlegendentry{thicket}
\addplot[cbase, densely dotted, line width=0.8pt] coordinates {(0.0008,6.3)(1.3,6.3)}; \addlegendentry{base}
\addplot[black, dashdotted, line width=0.8pt] coordinates {(0.0008,31.9)(1.3,31.9)}; \addlegendentry{majority class}
\end{axis}
\end{tikzpicture}
\text{\footnotesize \shortstack{(b) Weak-model \\ temperature sweep}}
\end{minipage}
\hfill
\begin{minipage}[t]{0.32\linewidth}
\centering
\tikzset{
  model7B/.style   ={cthicket, mark options={fill=cthicket, draw=cthicket!55!black, line width=0.4pt}},
  model3B/.style   ={cextra, mark options={fill=cextra, draw=cextra!55!black, line width=0.4pt}},
  model15B/.style  ={cttmv, mark options={fill=cttmv, draw=cttmv!55!black, line width=0.4pt}},
  model05B/.style  ={csoup, mark options={fill=csoup, draw=csoup!55!black, line width=0.4pt}},
  taskCD/.style    ={mark=*},
  taskGSM/.style   ={mark=diamond*},
  taskMATH/.style  ={mark=square*},
  taskOLY/.style   ={mark=triangle*},
  taskMBPP/.style  ={mark=pentagon*},
  taskROC/.style   ={mark=star},
  taskUSPTO/.style ={mark=x},
}%
\renewcommand{\legendmark}[2][]{%
  \tikz[baseline=-0.5ex]
    \draw[
      only marks,
      mark size=1.6pt,
      #1,
      #2
    ] plot coordinates {(0,0)};
}
\begin{tikzpicture}
\begin{axis}[
  paperaxis,
  width=\linewidth,
  height=4.4cm,
  xlabel={full-weight thicket (\%)},
  ylabel={TT-MV (\%)},
  xmin=0, xmax=95,
  ymin=0, ymax=95,
  xtick={20,50,80},
  ytick={20,50,80},
  clip=false,
]

\addplot[
  gray,
  dashed,
  line width=0.5pt,
  domain=0:95,
  forget plot
] {x};

\taskpoint{90.75}{92.05}{model7B}{taskCD}
\taskpoint{89.76}{89.39}{model7B}{taskGSM}
\taskpoint{74.67}{77.67}{model7B}{taskMATH}
\taskpoint{46.20}{49.79}{model7B}{taskOLY}
\taskpoint{76.44}{74.35}{model7B}{taskMBPP}
\taskpoint{65.57}{65.33}{model7B}{taskROC}
\taskpoint{39.36}{36.26}{model7B}{taskUSPTO}

\taskpoint{37.40}{24.90}{model15B}{taskCD}
\taskpoint{76.72}{79.38}{model15B}{taskGSM}
\taskpoint{57.67}{59.33}{model15B}{taskMATH}
\taskpoint{28.48}{29.32}{model15B}{taskOLY}
\taskpoint{52.49}{46.20}{model15B}{taskMBPP}
\taskpoint{48.98}{47.71}{model15B}{taskROC}
\taskpoint{24.88}{ 2.30}{model15B}{taskUSPTO}

\taskpoint{56.03}{57.39}{model05B}{taskGSM}
\taskpoint{39.00}{41.00}{model05B}{taskMATH}
\taskpoint{14.35}{15.40}{model05B}{taskOLY}
\taskpoint{28.80}{11.91}{model05B}{taskMBPP}
\taskpoint{31.85}{31.89}{model05B}{taskROC}
\node[
  anchor=south east,
  font=\tiny,
  inner sep=1pt,
  fill=white,
] at (axis description cs:0.98,0.04) {%
  \begin{tabular}{@{}c@{\hspace{2pt}}l@{}}
    \legendmark[model7B]{taskCD}   & 7B   \\
    \legendmark[model15B]{taskCD}  & 1.5B \\
    \legendmark[model05B]{taskCD}  & 0.5B
  \end{tabular}%
};

\node[
  anchor=north west,
  font=\tiny,
  inner sep=0pt,
  align=center,
] at (axis description cs:0.04,0.98) {%
  \begin{tabular}{@{}c@{\hspace{1pt}}l@{\hspace{5pt}}
                     c@{\hspace{1pt}}l@{\hspace{5pt}}
                  @{}}
    \legendmark[black]{taskCD}   & countdown \\
    \legendmark[black]{taskGSM}  & GSM8K \\
    \legendmark[black]{taskMATH} & MATH-500 \\
    \legendmark[black]{taskOLY}  & Olympiad \\
    \legendmark[black]{taskMBPP} & MBPP \\
    \legendmark[black]{taskROC}  & ROCS \\
    \legendmark[black]{taskUSPTO} & USPTO
  \end{tabular}%
};

\end{axis}
\end{tikzpicture}
\text{\footnotesize \shortstack{ (c) TT-MV vs.\\ full-weight thicket}}
\end{minipage}

%% file: tables/tab_master.tex
\centering
\small
\begin{tabular}{lccccccc}
\toprule
task & base & soup & thicket & TT-MV & logit PSV \\
\midrule
countdown      & 66.6 & 68.4 & \underline{91.7} & \textbf{{92.1}} & {90.7} \\
\rowcolor{gray!10} GSM8K          & 87.5 & 87.6& {\underline{89.8}} & {89.4} & \textbf{{91.9}} \\
MATH-500       & 63.3 & 62.0& \underline{80.7} & 77.7 & \textbf{{81.3}} \\
\rowcolor{gray!10} OlympiadBench  & 38.0 & 36.5 & 48.3 & \underline{49.8} & \textbf{{52.5}} \\
MBPP           & 64.3 & 66.1 & {\textbf{76.1}} & {\underline{74.4}} & {73.2} \\
\rowcolor{gray!10} ROCStories     & 65.2 & 65.3 & \underline{65.6} & 65.3 & \textbf{65.8} \\
USPTO         & 28.6 & 31.9 & \underline{42.6} & {36.3} & \textbf{{42.7}} \\
\bottomrule
\end{tabular}

%% file: tables/linear.tex
\begin{tabular}{lc}
        \toprule
        Perturbation scale $\sigma$ & Agreement \\
        \midrule
        $2.5\times 10^{-4}$ & $97.5 \pm 1.9\%$ \\
        $5\times 10^{-4}$   & $94.7 \pm 2.0\%$ \\
        $\mathbf{1\times 10^{-3}}$ & $\mathbf{89.7 \pm 2.1\%}$ \\
        $2\times 10^{-3}$   & $75.9 \pm 2.6\%$ \\
        $4\times 10^{-3}$   & $47.2 \pm 2.1\%$ \\
        $6\times 10^{-3}$   & $25.3 \pm 1.4\%$ \\
        \bottomrule
    \end{tabular}

%% file: sections/77_vision.tex
\section{Extending PSV to Vision}
\label{sec:vision}

Vision differs from language in two key ways that matter for our analysis. First, LLMs reuse the same token-prediction head learned during pretraining across downstream tasks.
In contrast, vision foundation models are typically adapted by fitting a new linear probe on the target-task training set. The vision probe can therefore absorb domain-specific calibration effects that remain exposed in the LLM output distribution. Second, language models generate autoregressive trajectories, so voting over sampled outputs can benefit from self-consistency~\citep{wang2023selfconsistency}. A vision classifier instead produces a single class prediction per image, with no analogous trajectory-level diversity.

Based on these differences, we transpose the Neural Thickets hypothesis to vision by changing the notion of specialization. We restrict our analysis to classification, but vary the image domains. We therefore define a vision \emph{domain expert} as a perturbed backbone whose frozen representations support a better linear probe on a given target domain. If such experts exist in the local weight neighborhood, perturbing the backbone and refitting the probe should improve transfer. Conversely, if the gains arise mainly from output-space effects, perturbing only the class logits should be sufficient. We test these alternatives with a self-supervised DINOv3 backbone~\citep{simeoni2025dinov3}, comparing full-weight PSV, patch-embedding PSV, and logit PSV.

\begin{table}[t] 
\centering 
\caption{
\textbf{Vision PSV under domain shift.}
We report improvements relative to linear probing on the frozen DINOv3-7B backbone using worst-region accuracy for FMoW, worst-group accuracy for Waterbirds, and balanced top-1 accuracy for ImageNet. PSV variants use validation-selected ensembles with $K=10$; closed-form denotes a single deterministic output recalibration. Weight-space perturbations yield little improvement, whereas output-space methods produce substantial gains on FMow, Waterbirds, which present strong train-to-test shift.}
\label{tab:vision} 
\input{tables/tab_vision}
\end{table}

\paragraph{Full-weight PSV for vision.}
Let $f\left(\cdot;\vtheta\right)$ be a pretrained vision backbone. A downstream classifier is obtained by fitting a linear probe on features $f\left(x;\vtheta\right)$ from a labeled target-domain training set. For each full-weight PSV member, we perturb the backbone,
\begin{align}
    \vtheta_i
    =
    \vtheta
    +
    \sigma\boldsymbol{\epsilon}_i,
    \qquad
    \boldsymbol{\epsilon}_i
    \sim
    \mathcal{N}
    \left(
        \mathbf{0},
        \mathbf{I}_{d_{\vtheta}}
    \right),
\end{align}
recompute the features, and fit a new linear probe. Members are scored on a validation set, the top $K$ are retained, and their predictions are aggregated at test time. This is our direct translation of Neural Thickets to vision: the perturbation acts on the pretrained backbone, where a neighborhood of domain-specialized representations would be expected to reside.

\paragraph{Patch PSV for vision.}
As an intermediate weight-space variant, we perturb only the patch-embedding module and refit the probe, keeping the rest of the backbone fixed. This tests whether input-level perturbations suffice for domain adaptation.

\paragraph{Logit PSV for vision.}
The output-space variant fits a single linear probe on the unperturbed frozen backbone, then keeps both backbone and probe fixed. Each PSV member is obtained by adding a Gaussian bias to the class logits, following Equation~\ref{eq:logitbias}, with $V$ replaced by the number of classes. The same validation-based selection and test-time voting procedure is applied. This variant changes neither the representation nor the probe and is substantially cheaper, since the probe is fitted only once rather than once per perturbation.

\paragraph{Closed-form recalibration.}
To test whether the logit-PSV gains can be explained by simple output recalibration {~\citep{saerens2002adjusting,menon2021longtail}}, we also consider a deterministic baseline. For each class $c$, we add the fixed bias
\begin{align}
    b_c
    =
    \log \pi_{\mathrm{true}}\left(c\right)
    -
    \log \pi_{\mathrm{pred}}\left(c\right),
\end{align}
where $\pi_{\mathrm{true}}\left(c\right)$ and $\pi_{\mathrm{pred}}\left(c\right)$ are respectively the true and predicted class frequencies on the validation group of interest. This correction shifts the predicted class marginal toward the observed validation marginal. For binary classification, it reduces to adjusting the decision threshold.

\paragraph{Results.}
Table~\ref{tab:vision} reports gains over the base linear probe. We consider two datasets exhibiting strong train-to-test distribution shifts for the probe: FMoW~\citep{christie2018functionalmapworld}, evaluated with worst-region accuracy, and Waterbirds~\citep{sagawa2020distributionallyrobustneuralnetworks}, evaluated with worst-group accuracy. We additionally include balanced ImageNet~\citep{Deng2009ImageNetAL} as a control setting without such a shift.
The two weight-space variants, full-weight and patch PSV, have little effect, improving worst-group accuracy by at most one point even for the $6.7$B backbone. By contrast, logit PSV improves worst-group accuracy by $5$-$6$ points while leaving both backbone and probe unchanged. Closed-form recalibration recovers most of this gain on FMoW; on the binary Waterbirds task, it matches logit PSV by effectively selecting an improved decision threshold. On balanced ImageNet, none of the methods provides a substantial gain.

\paragraph{Analysis.}
These results suggest that the vision gains are primarily a recalibration effect. Logit PSV improves worst-group performance by roughly $5$ to $6$ points on FMoW and Waterbirds, both of which exhibit substantial train-to-test shifts, but provides essentially no gain on balanced ImageNet, where train and evaluation distributions are aligned. The closed-form correction reinforces this interpretation, recovering most of the logit-PSV gain on the shifted datasets by directly recalibrating the class marginal, while again yielding no meaningful improvement on ImageNet.

Furthermore, if random backbone perturbations revealed domain-specialized representations, refitting a probe on them should improve target-domain performance. We do not observe this, even at the largest scale considered, $6.7$B parameters. Instead, gains appear only when the output distribution is modified, suggesting that vision PSV is better explained by recalibration under distribution shift than by nearby domain experts in weight space.

%% file: tables/tab_vision.tex
\begin{tabular}{l c c c c c}\toprule
 & DINOv3 & \multicolumn{2}{c}{weight-space PSV} & \multicolumn{2}{c}{output-space} \\
 \cmidrule(lr){3-4}\cmidrule(lr){5-6}
 Dataset (metric) & base & thicket & patch-PSV & logit PSV & closed-form \\\midrule
 ImageNet (bal.\ top-1)   & 88.4 & $0.0$   & \bf $+0.1$ & $-0.2$          & n/a   \\
 FMoW-OOD (worst-region)  & 35.1 & $+1.0$  & $+0.3$ & $\mathbf{+6.2}$ & $+3.7$ \\
 Waterbirds (worst-group) & 78.9 & $+0.3$  & $+0.0$ & $\mathbf{+5.0}$ & $+5.0$ \\
 \bottomrule
\end{tabular}

%% file: sections/7_discussion.tex
\section{Discussion}
\label{sec:discussion}

The gains of Neural Thickets do not require perturbing the full model: comparable improvements arise from perturbing a single decoder block or only the logits of a frozen model. This supports interpreting Neural Thickets through perturb-select-vote, connecting them to perturb-and-MAP sampling, self-consistency, and verifier-guided selection. In vision, gains under domain shift likewise arise primarily at the classifier output. These findings do not rule out useful local structure in weight space, but show that it is not required for the gains studied here.

\paragraph{Limitations.}
Our language experiments stop at 7B parameters; larger models may exhibit weight-space effects that logit perturbations fail to capture. PSV requires a labeled selection set and a way to compare generated answers for voting, limiting its applicability to open-ended generation. Our first-order analysis is valid only for sufficiently small $\sigma$. Finally, like full-weight PSV, logit PSV requires $K$ model evaluations per test input, increasing inference cost relative to a single prediction.

%% file: sections/9_statements.tex
\subsection*{AI use statement}
In this work, we used generative AI tools to assist with the writing of proofs, provide feedback on research methodology and experimental design, and support method implementation. We did not use generative AI tools for generating synthetic datasets, developing theoretical models or conceptual frameworks, formulating mathematical claims or providing key ingredients of proofs, proposing hypotheses, translation, dataset cleaning or reformatting, qualitative or thematic data analysis, or interpretation of results. Additionally, we used generative AI tools for scientific figure design, suggesting experimental parameters, creating and editing software code, brainstorming, drafting and editing parts of the manuscript for clarity and readability, identifying potentially relevant literature, and proposing or refining titles and keywords.

All AI-assisted outputs were reviewed and verified by the authors: code was tested for correctness, methodological and mathematical suggestions were checked independently, literature references were verified against the original sources, and text and figures were manually revised. The authors take full responsibility for the final content of the paper, including all claims, code, figures, and text produced with the assistance of generative AI.

\subsection*{Reproducibility statement}
All datasets and pretrained models used in this work are publicly available. The main paper and appendix document the experimental protocols, perturbation distributions, selection and voting procedures, hyperparameters, and evaluation settings used throughout the study. We will release the complete code for reproducing the LLM experiments upon publication.

\subsection*{Ethics statement}
This work analyzes and simplifies inference-time adaptation methods for pretrained models. All experiments use publicly available models and datasets; no new data were collected and no human subjects were involved. Logit PSV only reweights the outputs of a frozen model, so it inherits the biases and failure modes of the underlying model, and, like other training-free adaptation methods, it could in principle be used to steer a model toward undesirable behavior. We evaluate it only on standard reasoning, chemistry, and image recognition benchmarks, and we do not expect it to introduce risks beyond those of the pretrained models it builds on.

\subsection*{Acknowledgments}
We thank Mathieu Aubry and Alexei A. Efros for helpful discussions and advice, and Phillip Isola for feedback on this work. Loic Landrieu was supported by Hi!\ PARIS (ANR-23-IACL-0005) and ANR project sharp ANR-23-PEIA-0008 in the context of the PEPR IA. Experiments involving third-party models and datasets---namely OLMo-3-7B-Instruct, Qwen2.5-0.5B-Instruct, Qwen2.5-1.5B-Instruct, GSM8K, MATH-500, OlympiadBench, MBPP, ROCStories, and USPTO---were conducted without the involvement of Meta.

%% file: sections/8_appendix.tex
This appendix provides additional material supporting the main paper. We first present an extended discussion of related work (Section~\ref{app:extended_related}), followed by additional theoretical analysis (Section~\ref{app:proof}), cross-scale results and ablations (Section~\ref{app:cross}), and further experimental details (Section~\ref{app:experiment}).

\section{Extended Related work}
\label{app:extended_related}

\textbf{Test-time sampling and answer selection.}
A long line of work improves fixed models at inference by sampling multiple outputs and aggregating or selecting them. Self-consistency returns the majority answer over diverse reasoning paths~\citep{wang2023selfconsistency}; repeated sampling improves coverage as the sample count grows~\citep{brown2024monkeys}, and allocating inference compute effectively can match much larger models~\citep{snell2024testtime}. When a verifier or reward model is available, best-of-$N$ selects the highest-scoring candidate~\citep{cobbe2021verifiers}, while minimum Bayes risk decoding aggregates over the sampled distribution~\citep{eikema2020mbr}. Recent work further studies the coverage, optimality, and reward-model dependence of best-of-$N$ and related test-time scaling procedures~\citep{huang2025best,sriraman2026revisiting,li2026should}. Perturb-and-MAP provides the classical link between perturbation and sampling: adding Gumbel noise to logits and taking the arg-max yields exact categorical sampling~\citep{papandreou2011perturb,maddison2017concrete,jang2017gumbel}, closely related to temperature and nucleus sampling~\citep{holtzman2020nucleus}. We place Neural Thickets~\citep{gan2026thickets} in this broader family: perturbations generate diversity, validation-based selection favors high-reward members, and voting aggregates their predictions.

\textbf{Logit-space control.}
A growing body of work modifies model behavior directly in logit space while keeping the underlying predictor fixed. This includes watermarking, which biases selected token logits to embed a detectable signal~\citep{kirchenbauer2023watermark}, classifier-free guidance~\citep{ho2022cfg,sanchez2023cfg}, contrastive decoding~\citep{li2023contrastive}, decoding-time experts~\citep{liu2021dexperts}, and proxy-tuning~\citep{liu2024proxy}. Post-hoc calibration adjusts the outputs of a fixed model with a temperature~\citep{guo2017calibration} or with per-class corrections for new class priors~\citep{saerens2002adjusting}, long-tailed classes~\citep{menon2021longtail}, or the answer biases of language models~\citep{zhao2021calibrate}. Class-level logit perturbations have also been learned for single- and multi-label classification~\citep{li2023class}, sample-wise ones against spurious correlations~\citep{zhou2025mitigating}, and corpus-specific token scores steer frozen language models~\citep{an2026steering}. These methods design or fit the logit modification for a specific task. In contrast, logit PSV samples task-agnostic random biases.

\textbf{Merging and connectivity in weight space.}
A substantial literature studies when different solutions can be combined in weight space. Stochastic weight averaging averages checkpoints along a training trajectory to improve generalization~\citep{izmailov2018swa}, while model soups average independently fine-tuned models without sacrificing accuracy~\citep{wortsman2022soups}. Related work composes task vectors~\citep{ilharco2023task} or explicitly merges multiple specialists~\citep{matena2022merging,yadav2023ties}. These observations are closely connected to mode connectivity~\citep{garipov2018mode}, linear connectivity after accounting for permutation symmetries~\citep{frankle2020lmc,ainsworth2023rebasin}, and shared low-loss basins across tasks~\citep{mirzadeh2021lmc}.

\textbf{Low-dimensional structure and random perturbations.}
A substantial literature suggests that useful adaptation directions can occupy low-dimensional subspaces. Objective landscapes can have low intrinsic dimension~\citep{li2018intrinsic}, language-model fine-tuning often lies in low-dimensional subspaces~\citep{aghajanyan2021intrinsic}, and low-rank adapters can efficiently specialize large models~\citep{hu2022lora}. Relatedly, evolution strategies and zeroth-order methods optimize models by evaluating random weight perturbations~\citep{salimans2017es,malladi2023mezo}. Unlike Neural Thickets, however, these methods explicitly optimize the perturbations rather than selecting among unstructured random draws. Concurrent work similarly explores more structured searches of the local weight neighborhood, either by consolidating rewarded perturbations~\citep{zhang2026corp} or by using Bayesian optimization to identify strong single experts with fewer evaluations~\citep{cendra2026can}. These results are complementary to ours: structured weight-space search may be useful even if the gains of random PSV do not require a weight-space explanation.

\section{Persistent Perturbations, Selection, and Voting: Formal Statements}
\label{app:proof}

This section gives the formal statements underlying Section~\ref{sec:theory}. Since the vote operates on final answers, we work with distributions over complete answers rather than individual tokens. Section~\ref{app:setup} shows that logit PSV defines persistent vocabulary biases, Section~\ref{app:vote} characterizes voting over the resulting answer distribution, and Section~\ref{app:select} shows why selection requires perturbations that persist across inputs. Section~\ref{app:temperature} relates logit PSV to temperature sampling; the remaining results do not depend on this connection. The complete logit PSV procedure is given in Algorithm~\ref{alg:logitpsv}.

\begin{algorithm}[t]
\caption{Logit perturb-select-vote}
\label{alg:logitpsv}
\begin{algorithmic}[1]
\Require frozen model with logits $\boldsymbol{\ell}\left(\cdot\right)$, selection reward $R$, noise scale $\sigma$, population size $N$, ensemble size $K$
\For{$i = 1,\dots,N$}
    \State draw $\boldsymbol{\epsilon}_i \sim \mathcal{N}\left(\mathbf{0}, \mathbf{I}_{V}\right)$
    \State decode greedily under logits $\boldsymbol{\ell} + \sigma \boldsymbol{\epsilon}_i$, with $\boldsymbol{\epsilon}_i$ fixed across tokens {and inputs}
    \State $r_i \gets R\left(i\right)$ reward on the validation set
\EndFor
\State $\mathcal{S} \gets$ indices of the top $K$ members by $r_i$
\State \textbf{return} the majority vote over members $i \in \mathcal{S}$ at test time
\end{algorithmic}
\end{algorithm}

\subsection{Each member recalibrates the vocabulary}
\label{app:setup}

The model decodes an input $x$ autoregressively. At step $t$, it emits logits $\boldsymbol{\ell}_t = \boldsymbol{\ell}(x, y_{<t}) \in \mathbb{R}^{V}$ given the prefix $y_{<t}$, with next-token distribution $p(\cdot \mid x, y_{<t}) = \mathrm{softmax}(\boldsymbol{\ell}_t)$. Decoding stops at an end token or after $L$ steps, and a deterministic map $\mathrm{ans}$ extracts the final answer from the output sequence $y$. A logit-PSV member draws one bias $\sigma\boldsymbol{\epsilon} \in \mathbb{R}^{V}$ and decodes greedily,
\begin{equation}
    y_t
    =
    \operatorname*{arg\,max}_{v}
    \left(\ell_{t,v} + \sigma\epsilon_v\right),
    \label{eq:app-choice}
\end{equation}
with the same $\boldsymbol{\epsilon}$ at every step and on every input. Since $\mathrm{softmax}(\boldsymbol{\ell}_t + \sigma\boldsymbol{\epsilon})_v \propto p(v \mid x, y_{<t})\, e^{\sigma\epsilon_v}$, Equation~\ref{eq:app-choice} greedily decodes the recalibrated model of Equation~\ref{eq:recalib}, which multiplies the probability of each token by a fixed factor. This is the per-token counterpart of the per-class recalibration of Section~\ref{sec:vision}, with a random bias in place of a fitted one.

A member is thus a deterministic function of its bias. We write $y^{\boldsymbol{\epsilon}}(x)$ for its output and $a_{\boldsymbol{\epsilon}}(x) = \mathrm{ans}(y^{\boldsymbol{\epsilon}}(x))$ for its answer. Full-weight PSV fits the same setup, with $\boldsymbol{\epsilon}$ a weight perturbation and $y^{\boldsymbol{\epsilon}}(x)$ the greedy output of the perturbed network. In both variants, the answers of the members follow the answer distribution
\begin{equation}
    q_{\sigma}\left(a \mid x\right)
    =
    \mathrm{Pr}_{\boldsymbol{\epsilon}}
    \left[
        a_{\boldsymbol{\epsilon}}\left(x\right) = a
    \right],
    \label{eq:answer_law}
\end{equation}
where the probability is over the draw of the perturbation.

\subsection{Voting returns the most frequent answer}
\label{app:vote}

\begin{proposition}[Voting over random members]
\label{prop:vote}
Fix an input $x$ and draw $K$ members independently, without selection. Their answers $a_1, \dots, a_K$ are independent, with law $q_\sigma(\cdot \mid x)$. Let $\mathcal{A}_x$ be the set of answers with positive probability, assume that the most frequent answer $a^\star = \operatorname*{arg\,max}_a q_\sigma(a \mid x)$ is unique, and let $\gamma = q_\sigma(a^\star \mid x) - \max_{a \neq a^\star} q_\sigma(a \mid x) > 0$ be its margin. The plurality vote then returns $a^\star$ with probability at least $1 - \left(|\mathcal{A}_x| - 1\right) e^{-K\gamma^2/2}$.
\end{proposition}

\begin{proof}
Each answer $a_i = a_{\boldsymbol{\epsilon}_i}(x)$ is a fixed function of an independent draw $\boldsymbol{\epsilon}_i$, so the answers are independent with law $q_\sigma(\cdot \mid x)$. Outputs have at most $L$ tokens from a finite vocabulary, so $\mathcal{A}_x$ is finite. For $a \neq a^\star$, let $D_a = \sum_{i=1}^{K} \left(\mathbf{1}[a_i = a^\star] - \mathbf{1}[a_i = a]\right)$. Its $K$ terms are independent, take values in $[-1, 1]$, and have mean at least $\gamma$. Hoeffding's inequality gives $\Pr[D_a \le 0] \le e^{-K\gamma^2/2}$. The vote returns $a^\star$ whenever $D_a > 0$ for every $a \neq a^\star$, and a union bound over the $|\mathcal{A}_x| - 1$ competitors concludes.
\end{proof}

The same proof applies to TT-MV, with the answer distribution $p_T$ of temperature sampling (Equation~\ref{eq:app-pT}) in place of $q_\sigma$, which is self-consistency~\citep{wang2023selfconsistency}. Thus, as the number of voters grows, the unselected PSV vote converges to the most frequent answer under $q_\sigma(\cdot\mid x)$. It is therefore asymptotically correct on $x$ when this modal answer is correct. The same interpretation applies to TT-MV with $p_T(\cdot\mid x)$.

\subsection{Selection exploits persistence across inputs}
\label{app:select}

A member applies the same recalibration on the validation set and on test inputs. Selection ranks members by their reward $R_i$ on the validation set and keeps the top $K$. Its effect on a test input $x$ depends on whether $R_i$ carries information about the test answer $a_i(x)$.

\begin{proposition}[Selection needs members that persist across inputs]
\label{prop:persist}
Let the pairs $(R_i, a_i(x))$, for $i = 1, \dots, N$, be independent and identically distributed, and let $\mathcal{S}$ contain the $K$ members with the highest rewards, with ties broken by independent randomness. If $R_i$ is independent of $a_i(x)$, the selected answers $(a_i(x))_{i \in \mathcal{S}}$ are independent, with the same law as $a_1(x)$. The selected vote then has the same distribution as a vote over $K$ unselected members.
\end{proposition}

\begin{proof}
The pairs are independent across members, and each pair has independent components. The reward vector $(R_1, \dots, R_N)$ is therefore independent of the answer vector $(a_1(x), \dots, a_N(x))$. The set $\mathcal{S}$ depends only on the rewards and on the tie-breaking randomness, so it is independent of the answers. Conditioned on $\mathcal{S}$, the selected answers are thus still independent draws from the law of $a_1(x)$.
\end{proof}

This independence condition holds if the perturbation is resampled independently for every input, as in ordinary temperature sampling. If one were to rank such samples using rewards obtained on the selection set and then resample them on a test input, the selected indices would carry no information about their test answers. Selection therefore cannot change the test-answer distribution. In PSV, by contrast, each member retains the same perturbation across the selection and test inputs. The reward $R(\boldsymbol{\epsilon})$ and the answer $a_{\boldsymbol{\epsilon}}(x)$ then depend on the same $\boldsymbol{\epsilon}$, and selection can change the answer distribution. The next statement describes this change for a large population.

\begin{proposition}[Selection limit]
\label{prop:select}
Draw $\boldsymbol{\epsilon}^{(1)}, \dots, \boldsymbol{\epsilon}^{(N)}$ independently from the bias prior $p$, and let $R(\boldsymbol{\epsilon})$ be the member validation reward, with a continuous distribution. Keep the top $K = \lceil qN \rceil$ members by reward for a fixed fraction $q \in (0,1)$, and let $\tau_q$ be the unique level with $\Pr[R(\boldsymbol{\epsilon}) \ge \tau_q] = q$. Then as $N \to \infty$ the empirical distribution of the kept members converges weakly to $p_{\mathrm{sel}}$ with density
\begin{equation*}
p_{\mathrm{sel}}(\boldsymbol{\epsilon}) \;\propto\; p(\boldsymbol{\epsilon})\, \mathbf{1}\!\left[\, R(\boldsymbol{\epsilon}) \ge \tau_q \,\right].
\end{equation*}
\end{proposition}

\begin{proof}
Let $\hat{\tau}_N$ be the $K$-th largest reward. The Glivenko--Cantelli theorem, the continuity of the reward law, and the uniqueness of $\tau_q$ give $\hat{\tau}_N \to \tau_q$ almost surely. For a bounded continuous function $h$, the kept members satisfy
\begin{equation*}
    \frac{1}{K}\sum_{i \in \mathcal{S}} h\left(\boldsymbol{\epsilon}^{(i)}\right)
    =
    \frac{N}{K}\cdot\frac{1}{N}\sum_{i=1}^{N} h\left(\boldsymbol{\epsilon}^{(i)}\right)\mathbf{1}\left[R\left(\boldsymbol{\epsilon}^{(i)}\right) \ge \hat{\tau}_N\right].
\end{equation*}
Replacing $\hat{\tau}_N$ by $\tau_q$ changes the right-hand average by at most $\|h\|_\infty$ times the fraction of rewards between the two levels, which vanishes almost surely by the same theorem. The strong law of large numbers and $N/K \to 1/q$ then give the limit $\mathbb{E}[h(\boldsymbol{\epsilon})\mathbf{1}[R(\boldsymbol{\epsilon}) \ge \tau_q]]/q = \mathbb{E}_{p_{\mathrm{sel}}}[h]$.
\end{proof}

The continuity assumption is an idealization. A reward averaged over $200$ examples takes finitely many values, and ties at the threshold are then split at random. The experiments also fix $K = 50$ rather than the fraction $q$.

\begin{remark}[When selection helps]
\label{rem:weak}
With many selected members, the vote on $x$ converges to the most frequent answer under the selected distribution
$q_{\mathrm{sel}}(a \mid x)
=
\mathrm{Pr}_{\boldsymbol{\epsilon} \sim p_{\mathrm{sel}}}
\left[
a_{\boldsymbol{\epsilon}}(x)=a
\right]$,
rather than under $q_\sigma(\cdot \mid x)$. Selection can therefore help only insofar as the selection reward is informative about test behavior across members. In particular, it can change an incorrect plurality into a correct one, but can also do the reverse. This is not guaranteed: the selection set contains only $200$ examples, and individual members can overfit it (Appendix~\ref{app:cross}, geometry of the selected perturbations). Selection has the most room to help when the unselected answer distribution is unreliable, as in the weak-model regime of Figure~\ref{fig:ttmv}(b).
\end{remark}

\subsection{Relation to temperature sampling}
\label{app:temperature}

Logit PSV and temperature sampling both inject randomness at the output, but differ in how this randomness persists during decoding. Temperature-$T$ sampling independently resamples the perturbation at every decoding step, whereas logit PSV draws a single perturbation that is reused throughout decoding and across inputs. We refer to these two cases as \emph{resampled} and \emph{persistent} perturbations, respectively:
\begin{align}
    p_T\left(a \mid x\right)
    &=
    \sum_{y \,:\, \mathrm{ans}\left(y\right)=a}
    \prod_t
    \mathrm{softmax}
    \left(
        \boldsymbol{\ell}\left(x,y_{<t}\right)/T
    \right)_{y_t}.
    \label{eq:app-pT}
\end{align}
We compare the \emph{persistent perturbation} of logit PSV, where one $\boldsymbol{\epsilon}$ is reused at every step and on every input, with \emph{resampled perturbations}, where an independent vector $\boldsymbol{\epsilon}_t$ is drawn at every step and on every input.

\subsubsection{Resampled Gumbel perturbations recover temperature sampling}
\label{app:gumbel}

\begin{lemma}[Gumbel-max]
\label{lem:gumbel}
Let $g_v \sim \mathrm{Gumbel}(0,1)$ independently, with cumulative distribution $F(x) = e^{-e^{-x}}$ and density $f(x) = e^{-x}e^{-e^{-x}}$. Then for every $k$,
\begin{equation*}
\Pr\!\left[\operatorname*{arg\,max}_{v}\,(\ell_v + g_v) = k\right] = \frac{e^{\ell_k}}{\sum_{j=1}^{V} e^{\ell_j}} = \mathrm{softmax}(\boldsymbol{\ell})_k.
\end{equation*}
\end{lemma}

\begin{proof}
Condition on $g_k = z$. Token $k$ wins when $g_v \le \ell_k - \ell_v + z$ for all $v \neq k$, an event of probability $\prod_{v \neq k} F(\ell_k - \ell_v + z)$ by independence. Integrating over $z$,
\begin{equation*}
\Pr[\,y = k\,] = \int_{-\infty}^{\infty} f(z) \prod_{v \neq k} F(\ell_k - \ell_v + z)\, dz
= \int_{-\infty}^{\infty} e^{-z}e^{-e^{-z}} \exp\!\Big(-e^{-z}\, e^{-\ell_k}\!\!\sum_{v \neq k} e^{\ell_v}\Big) dz,
\end{equation*}
where the product collapses because $\prod_{v\neq k} \exp(-e^{-(\ell_k - \ell_v + z)}) = \exp(-e^{-z}e^{-\ell_k}\sum_{v\neq k}e^{\ell_v})$. Combining the two exponentials and writing $Z = \sum_{j} e^{\ell_j}$ gives the integrand $e^{-z}\exp(-e^{-z} e^{-\ell_k} Z)$. Substitute $u = e^{-z}$, so $e^{-z}\,dz = -du$ and the limits flip to $[0,\infty)$,
\begin{equation*}
\Pr[\,y = k\,] = \int_{0}^{\infty} \exp\!\big(-u\, e^{-\ell_k} Z\big)\, du = \frac{1}{e^{-\ell_k} Z} = \frac{e^{\ell_k}}{Z}. \qedhere
\end{equation*}
\end{proof}

\begin{corollary}[The noise scale is the temperature]
\label{cor:temp}
For $g_v \sim \mathrm{Gumbel}(0,1)$ and scale $\sigma > 0$, the token $\operatorname*{arg\,max}_v(\ell_v + \sigma g_v)$ has law $\mathrm{softmax}(\boldsymbol{\ell}/\sigma)$. Hence, at a single decoding step, the bias scale $\sigma$ equals the sampling temperature $T$.
\end{corollary}

\begin{proof}
The arg-max is invariant to a positive rescaling of its argument, so $\operatorname*{arg\,max}_v(\ell_v + \sigma g_v) = \operatorname*{arg\,max}_v(\ell_v/\sigma + g_v)$. Apply Lemma~\ref{lem:gumbel} with logits $\boldsymbol{\ell}/\sigma$.
\end{proof}

\begin{proposition}[Fresh Gumbel noise samples whole sequences]
\label{prop:ancestral}
Under fresh noise with independent $\mathrm{Gumbel}(0,1)$ entries, the decoded sequence has law $\prod_{t} \mathrm{softmax}(\boldsymbol{\ell}_t/\sigma)_{y_t}$, and its answer has law $p_{T=\sigma}(\cdot \mid x)$.
\end{proposition}

\begin{proof}
The prefix $y_{<t}$ is a function of $\boldsymbol{\epsilon}_1, \dots, \boldsymbol{\epsilon}_{t-1}$, which are independent of $\boldsymbol{\epsilon}_t$. Conditioned on $y_{<t}$, Corollary~\ref{cor:temp} with logits $\boldsymbol{\ell}_t$ thus gives $\Pr[y_t = k \mid y_{<t}] = \mathrm{softmax}(\boldsymbol{\ell}_t/\sigma)_k$. The chain rule gives the law of the sequence. The answer is a deterministic function of the sequence, so its law is $p_{T=\sigma}(\cdot \mid x)$ by Equation~\ref{eq:app-pT}.
\end{proof}

{With resampled Gumbel perturbations, voting over $K$ decoded sequences is therefore exactly TT-MV at temperature $T=\sigma$, and selection has no effect (Proposition~\ref{prop:persist}).}

\subsubsection{Persistent perturbations introduce memory across decisions}
\label{app:fixed}

A persistent perturbation has the same conditional effect as a resampled perturbation whenever the next decision depends on bias coordinates that have not influenced earlier decisions. We make this precise in an idealized restricted-candidate setting. Let $C_t(h)$ be the set of candidate tokens at step $t$ after the prefix $h = y_{<t}$, with logits set to $-\infty$ outside $C_t(h)$, and let $U(h) = \bigcup_{s<t} C_s(h_{<s})$ collect the tokens that competed at earlier steps.

\begin{proposition}[Disjoint decisions match resampled perturbations]
\label{prop:memory}
Consider a persistent perturbation with independent entries and a prefix $h$ with $\Pr[y_{<t} = h] > 0$. If $C_t(h) \cap U(h) = \emptyset$, the next token has the same conditional law as under a resampled perturbation. With Gumbel perturbations, this law is $\mathrm{softmax}(\boldsymbol{\ell}_t/\sigma)$ restricted to $C_t(h)$. If this condition holds at every step and for every prefix, persistent and resampled perturbations induce the same distributions over sequences and answers.
\end{proposition}

\begin{proof}
The event $\{y_{<t} = h\}$ requires that, at each step $s < t$, the token $h_s$ wins the arg-max over $C_s(h_{<s})$. This event therefore depends only on the entries $\epsilon_v$ with $v \in U(h)$, whereas the next token depends only on those with $v \in C_t(h)$. When the two sets are disjoint, independence implies that conditioning on $\{y_{<t} = h\}$ leaves the entries in $C_t(h)$ with their prior law, exactly as under resampling. Lemma~\ref{lem:gumbel}, applied to the logits $\boldsymbol{\ell}_t/\sigma$ restricted to $C_t(h)$, gives the Gumbel case. The final claim follows from the chain rule, as in Proposition~\ref{prop:ancestral}.
\end{proof}

For the untruncated decoder used by logit PSV, $C_t(h)$ is formally the full vocabulary, so the disjointness condition does not hold beyond the first step. Nevertheless, successive decisions may effectively depend on different subsets of competitive tokens, in which case earlier choices reveal little information about the bias coordinates relevant to the next one. Proposition~\ref{prop:memory} should therefore be viewed as an exact characterization of the disjoint case, rather than as an approximation guarantee for the full decoder.

\textbf{Memory.} When the same tokens compete again, the prefix carries information about their entries of the bias. The next token then follows the per-step law under the posterior of $\boldsymbol{\epsilon}$ given the prefix, instead of its prior:
\begin{align}
    \Pr\left[y_t = k \mid y_{<t}\right]
    &=
    \mathrm{Pr}_{\boldsymbol{\epsilon}}
    \left[
        \operatorname*{arg\,max}_{v}
        \left(
            \ell_{t,v} + \sigma\epsilon_v
        \right)
        = k
        \,\middle|\,
        y^{\boldsymbol{\epsilon}}_{<t} = y_{<t}
    \right].
    \label{eq:app-posterior}
\end{align}
The persistent perturbation therefore couples decisions whenever the same bias coordinates matter repeatedly, both within a sequence and across inputs. Across inputs, this is precisely the persistence that makes selection informative in Section~\ref{app:select}.

\textbf{Example: the same two tokens compete twice.} Consider two tokens $A$ and $B$, logits $(\delta, 0)$ at the first step and $(0, \delta)$ at the second, with $\delta > 0$, Gumbel noise, and $\sigma = 1$. The answer is the sequence itself, and the correct answer is $AB$. Let $p = 1/(1 + e^{-\delta}) > 1/2$. Resampled perturbations make the two choices independently, so $\Pr[AB] = p^2$, and $AB$ is the most frequent answer. With a persistent perturbation, both choices depend on the same difference $D = g_A - g_B$, which follows a standard logistic law. The member outputs $AB$ when $-\delta < D < \delta$, $AA$ when $D \ge \delta$, and $BB$ when $D \le -\delta$, so
\begin{equation}
    \Pr[AB] = 2p - 1,
    \qquad
    \Pr[AA] = \Pr[BB] = 1 - p,
    \qquad
    \Pr[BA] = 0.
    \label{eq:app-example}
\end{equation}
With $\delta = 0.4$, resampled perturbations give $\Pr[AB] = 0.36$, whereas the persistent perturbation gives $\Pr[AB] = 0.20$ and $\Pr[AA] = \Pr[BB] = 0.40$, so the asymptotic vote is incorrect. The effect reverses when the correct answer repeats one choice. With logits $(\delta, 0)$ at both steps, the fixed bias gives $\Pr[AA] = p$, against $p^2$ under fresh noise. This example illustrates the effect of persistence: reusing the same perturbation correlates repeated decisions. Depending on the structure of the correct sequence, this correlation can either increase or decrease its probability.

\subsubsection{Gaussian perturbations have a local temperature interpretation}
\label{app:gauss}

Logit PSV uses Gaussian rather than Gumbel perturbations. The statements below characterize the resulting law of a single decision. By Proposition~\ref{prop:memory}, the same law applies under a persistent perturbation whenever the relevant bias coordinates have not influenced earlier decisions.

\begin{lemma}[Gaussian perturb-and-MAP limits]
\label{lem:gauss}
Let $\epsilon_v \sim \mathcal{N}(0,1)$ independently and $y_\sigma = \operatorname*{arg\,max}_v(\ell_v + \sigma\epsilon_v)$, and suppose $\boldsymbol{\ell}$ has a unique maximizer $k^\star$. Then
\begin{align}
\Pr[\,y_\sigma = k^\star\,] \xrightarrow[\sigma \to 0]{} 1, \qquad\text{and}\qquad \Pr[\,y_\sigma = k\,] \xrightarrow[\sigma \to \infty]{} \tfrac{1}{V} \ \text{ for every } k.
\end{align}
\end{lemma}

\begin{proof}
For the first limit, write $\Delta_v = \ell_{k^\star} - \ell_v > 0$ for $v \neq k^\star$. A union bound gives
\begin{equation*}
\Pr[\,y_\sigma \neq k^\star\,] \le \sum_{v \neq k^\star} \Pr\!\left[\sigma(\epsilon_v - \epsilon_{k^\star}) > \Delta_v\right] = \sum_{v \neq k^\star} \overline{\Phi}\!\left(\frac{\Delta_v}{\sigma\sqrt{2}}\right),
\end{equation*}
since $\epsilon_v - \epsilon_{k^\star} \sim \mathcal{N}(0,2)$, where $\overline{\Phi}$ is the standard normal tail. Each argument diverges as $\sigma \to 0$, so each term and the finite sum vanish. For the second limit, $\operatorname*{arg\,max}_v(\ell_v + \sigma\epsilon_v) = \operatorname*{arg\,max}_v(\ell_v/\sigma + \epsilon_v)$, and $\ell_v/\sigma \to 0$ uniformly, so the maximizer converges to $\operatorname*{arg\,max}_v \epsilon_v$. The $\epsilon_v$ are independent and identically distributed with a continuous law, so this maximizer is uniform on $\{1,\dots,V\}$ by exchangeability.
\end{proof}

\begin{remark}
\label{rem:tails}
The exact choice probability is the multinomial probit integral
\begin{equation*}
\Pr[\,y_\sigma = k\,] = \int_{-\infty}^{\infty} \varphi(u) \prod_{v \neq k} \Phi\!\left(u + \frac{\ell_k - \ell_v}{\sigma}\right) du,
\end{equation*}
with $\varphi, \Phi$ the standard normal density and distribution. It has no softmax closed form, but it is continuous in $\sigma$ and connects the point mass at $k^\star$ (Lemma~\ref{lem:gauss}, $\sigma \to 0$) to the uniform law ($\sigma \to \infty$).
The scale $\sigma$ therefore controls the stochasticity of a single decision, analogously to a sampling temperature. This analogy can be made explicit for two competing tokens whose logits differ by $\Delta>0$. Gaussian perturbations select the lower-logit token with probability
$\overline{\Phi}\left(\Delta/(\sigma\sqrt{2})\right)$. Using the approximation
$\Phi(z) \approx 1/\left(1+e^{-1.702z}\right)$ gives
$1/\left(1+e^{\Delta/(0.83\sigma)}\right)$, corresponding locally to softmax sampling at temperature $T \approx 0.83\sigma$. The analogy breaks down for large logit gaps: the Gaussian probability is bounded by
$\tfrac{1}{2}e^{-\Delta^2/(4\sigma^2)}$, whereas the corresponding softmax probability decays as $e^{-\Delta/T}$. Gaussian perturbations therefore behave similarly to temperature sampling for near-tied alternatives, while suppressing large-logit-gap alternatives more strongly.
\end{remark}

\subsection{Internal perturbations act as logit perturbations}
\label{app:residual}

The localized perturbation results of Section~\ref{sec:local} can be understood through their effect on the output logits.

\begin{remark}
\label{rem:residual}
Let $\mathbf{h} \in \mathbb{R}^{d}$ be the residual stream at the decoding position, read out into logits by $\boldsymbol{\ell} = \mathbf{W}_U\,\mathrm{LN}(\mathbf{h})$ with unembedding $\mathbf{W}_U \in \mathbb{R}^{V \times d}$. A write matrix $\mathbf{W}$ (an attention or feed-forward output projection) contributes additively to the stream, $\mathbf{h} = \mathbf{h}_0 + \mathbf{W}\mathbf{a}$ for the sublayer activation $\mathbf{a}$. Perturbing it by $\mathbf{W} \mapsto \mathbf{W} + \boldsymbol{\Delta}$ shifts the stream by $\boldsymbol{\Delta}\mathbf{a}$ and, to first order, the logits by
\begin{align*}
    \delta\boldsymbol{\ell}
    &\approx
    \mathbf{W}_U
    \mathbf{J}_{\mathrm{LN}}(\mathbf{h})
    \boldsymbol{\Delta}\mathbf{a}.
\end{align*}
where $\mathbf{J}_{\mathrm{LN}}$ is the Jacobian of the LayerNorm. This is an additive perturbation of the output logits, as in Equation~\ref{eq:logitbias}, although its structure is inherited from the perturbed weights and the intervening network. Logit PSV corresponds to injecting such a shift directly at the output, without propagating a perturbation through the network. The same linearization applies to other parameters within the block. Let $\mathbf{W}$ act on an input $\mathbf{x}$ inside a sublayer whose contribution to the residual stream is $g\left(\mathbf{W}\mathbf{x}\right)$, where $g$ contains the subsequent frozen operations. Perturbing $\mathbf{W} \mapsto \mathbf{W} + \boldsymbol{\Delta}$ changes this contribution to first order by $\mathbf{J}_g\boldsymbol{\Delta}\mathbf{x}$, and hence changes the logits by: 
\begin{align*}
    \delta\boldsymbol{\ell}
    &\approx
    \mathbf{W}_U
    \mathbf{J}_{\mathrm{LN}}(\mathbf{h})
    \mathbf{J}_g
    \boldsymbol{\Delta}\mathbf{x}.
\end{align*}
Unless the perturbation lies in the null space of the resulting linear map, it therefore induces a first-order change in the logits. The location of the perturbation determines the structure and magnitude of this change through the intervening Jacobians. This provides a local explanation for why perturbations applied to very different components can ultimately act through the same output space.
For query and key projections, the perturbation additionally passes through the attention softmax, whose Jacobian
$\mathrm{diag}\left(\mathbf{p}\right)-\mathbf{p}\mathbf{p}^{\top}$
can attenuate changes before they reach the residual stream. This is consistent with Figure~\ref{fig:block}, where perturbing the query/key path recovers $85$--$89\%$ of the full-weight gain, compared with $96$--$103\%$ for the other components. We interpret this ordering as evidence that the intervening computation modulates the induced logit perturbation, rather than preventing it altogether.
\end{remark}

Together, these results describe PSV in output space: perturbations induce persistent changes in the answer distribution, selection favors perturbations that perform well on the selection set, and voting aggregates their answers. Remark~\ref{rem:residual} connects internal perturbations to this view through the logit shifts they induce.

\begin{table}[t]
\centering
\begin{minipage}[t]{0.5\linewidth}
\vspace{0pt}
\caption{\textbf{Accuracy of the first-order approximation.}
Fraction of perturbation-induced arg-max changes reproduced by the linearized logits on OLMo-3-7B-Instruct for Countdown. Bold indicates the perturbation scale used in our experiments.}
\label{tab:lin}
\end{minipage}
\hfill
\begin{minipage}[t]{0.45\linewidth}
\vspace{5pt}
\centering
\small
\begin{tabular}{rl}
\toprule
$\sigma$ & Agreement \\
\midrule
$2.5\times10^{-4}$ & $97.5 \pm 1.9\%$ \\
$5\times10^{-4}$   & $94.7 \pm 2.0\%$ \\
$\bm{1\times10^{-3}}$ & $\bm{89.7 \pm 2.1\%}$ \\
$2\times10^{-3}$   & $75.9 \pm 2.6\%$ \\
$4\times10^{-3}$   & $47.2 \pm 2.1\%$ \\
$6\times10^{-3}$   & $25.3 \pm 1.4\%$ \\
\bottomrule
\end{tabular}
\end{minipage}
\end{table}

\paragraph{Validating the linear approximation.}
Remark~\ref{rem:residual} treats the induced logit shift as locally linear in the weight perturbation. We measure the accuracy of this approximation on OLMo-3-7B-Instruct on Countdown. 
For a fixed teacher-forced prefix, let
$\mathbf{J}=\partial\boldsymbol{\ell}_{\boldsymbol{\theta}}/\partial\boldsymbol{\theta}$
denote the logit Jacobian at the pretrained parameters. The first-order prediction for a
perturbed model $\bm\theta+\sigma\bm\epsilon$ is
\begin{align}    \boldsymbol{\ell}^{\mathrm{lin}}_{\sigma,\boldsymbol{\epsilon}}
    &=
    \boldsymbol{\ell}_{\boldsymbol{\theta}}
    +
    \sigma\mathbf{J}\boldsymbol{\epsilon}.
\end{align}
We compute $\mathbf{J}\bm\epsilon$ exactly with a forward-mode Jacobian--vector
product and compare this prediction against the perturbed model's actual logits.
Both are evaluated along fixed token sequences through teacher forcing: each prompt from the selection set followed by the base model's greedy continuation. Each model has an arg-max token at every position and we report, over the positions where the perturbation changes that token, the fraction where the linearized effect changes it to the same one. We draw $32$ perturbations $\bm\epsilon$ and report $95\%$ confidence
intervals in Table~\ref{tab:lin}.

The first-order term alone accounts for
$89.7\%$ of the greedy choices a perturbation changes. Agreement falls off as the perturbation grows
(Table~\ref{tab:lin}), so this describes the regime the method actually uses
rather than holding everywhere. At the perturbation scale used in our experiments, the first-order approximation reproduces $89.7\%$ of the arg-max changes induced by the perturbed model. Agreement decreases as the perturbation scale grows (Table~\ref{tab:lin}), confirming that the approximation is local. These results support modeling the effect of the weight perturbation through its induced first-order logit shift in the regime used by PSV. They do not establish equivalence at the sequence level: the comparison is performed at fixed teacher-forced prefixes, whereas autoregressive decoding can amplify discrepancies across subsequent tokens.

\section{Cross-Scale Numbers and Ablations}
\label{app:cross}

Section~\ref{sec:ttmv} reports OLMo-3-7B-Instruct in the main table. The cross-scale comparisons behind Figures~\ref{fig:logitbias} and~\ref{fig:ttmv} are collected in Table~\ref{tab:crossscale}, and the layer ablations (perturbing all but the last block, the last block only, and the logit bias) are available in the released results table. 
{With its noise scale selected on the $200$ selection examples (the scale whose top-$50$ members have the highest mean training reward, among $\sigma \in \{0.002, 0.004, 0.008, 0.032\}$), the last-block-only perturbation matches or exceeds the full-weight result on the four reasoning tasks at 7B ($+0.9$ to $+2.5$ points) but trails it by $1.3$ to $2.2$ points on MBPP, ROCStories, and USPTO. On the Qwen models it ranges from $3.8$ points below the full-weight result (Qwen2.5-0.5B, MBPP) to $3.0$ points above it (Qwen2.5-1.5B, GSM8K). The selected scale tends to increase with model size.} We exclude the two degenerate cells (Qwen2.5-0.5B on Countdown and USPTO, where the base model cannot perform the task) from all conclusions.

\begin{table}[t]
\caption{\textbf{Cross-scale comparison of full-weight and output-space perturbations.}
Accuracy at $K{=}50$ (\%) for the greedy base, full-weight PSV, logit PSV, and an input-embedding-bias variant on Qwen2.5 and OLMo-3-7B-Instruct. The input-embedding variant adds a Gaussian bias to token embeddings and is evaluated on Qwen2.5 only. {The logit-bias scale is fixed per model rather than chosen per task ($\sigma{=}0.5$ for Qwen2.5-0.5B, $1.0$ for 1.5B and 3B, $2.0$ for OLMo-3-7B), and full-weight PSV uses $\sigma{=}0.001$. Every method uses $N{=}1000$, except full-weight PSV on Countdown ($N{=}2000$).} Degenerate Qwen2.5-0.5B cells on Countdown and USPTO are omitted. {$^{\dagger}$A constant predictor of the majority reaction class scores $31.9\%$ on USPTO; these cells are at or below it and are not bolded. $^{\ddagger}$Above that baseline, but with lower balanced accuracy than full-weight PSV ($26.7\%$ against $29.7\%$).}}
\input{tables/tab_crossscale}
\end{table}

Figure~\ref{fig:kcurve-qwen} traces the full $K$-sweep behind these numbers for the full-weight thicket and the logit-bias ensemble at all three Qwen scales, where the two track each other and the gain grows with model size.

\begin{figure}[t]
\input{figures/fig_kcurve_qwen}
\caption{\textbf{Performance as a function of ensemble size.}
Gain over the greedy base, in accuracy points, as a function of $K$ for full-weight PSV and logit PSV on Qwen2.5-0.5B, 1.5B, and 3B-Instruct ($N{=}1000$, {$N{=}2000$ for full-weight Countdown; $\sigma$ fixed per model as in Table~\ref{tab:crossscale}}). Countdown and USPTO are omitted at 0.5B because the base model cannot perform these tasks.%
}
\label{fig:kcurve-qwen}
\end{figure}

\paragraph{The mechanism holds across architectures.} Figure~\ref{fig:block-qwen} repeats the single-weight ablation of Section~\ref{sec:local} on Qwen2.5-0.5B, which differs from OLMo in three ways relevant here: it has no QK-norm, it adds biases to the query, key, and value projections, and it is pre-norm rather than post-norm. The conclusion is unchanged: perturbing any single component, including the newly present QKV biases, improves over base. The ordering is also consistent with the first-order account of Appendix~\ref{app:residual}. The weights that reach the stream through a linear or elementwise path are the most potent: at 0.5B the value and up projections ($+7.6$, $+7.1$) even exceed the output projections ($+5.5$). The weights that act only through the attention scores or through a normalized sublayer input are the weakest: the query and key projections ($+1.4$), the query and key biases ($+1.3$, $+1.1$), and the pre-norm gains ($+1.4$, $+1.7$). The pre-norms differ from the post-norms of OLMo, which scale the sublayer output written into the stream and match $\mathtt{o\_proj}$. On Countdown, the gain from every component increases with model scale (Figure~\ref{fig:block-qwen-scale}): at 1.5B the weakest components gain $+11$ to $+16$ and the value and output projections gain $+26$ to $+27$, above the full-block perturbation; at 3B every component gains between $+27.5$ and $+39.9$. Table~\ref{tab:component-scale} reports the per-component gains and training-reward spreads at the 1.5B and 3B scales.

\begin{figure}[t]
\input{figures/fig_block_qwen}
\caption{\textbf{Single-component perturbations transfer across architectures.}
Gain over base, in accuracy points, when perturbing one component of the first decoder block of Qwen2.5-0.5B-Instruct on GSM8K ($K{=}50$; base $44.3\%$, full-block gain $+11.8$). Every component improves over base, with the largest gains from the value, up, and output projections. Qwen2.5 uses pre-norm blocks and includes QKV biases, unlike OLMo.}
\label{fig:block-qwen}
\end{figure}

\begin{figure}[t]
\input{figures/fig_block_qwen_scale}
\caption{\textbf{Single-component perturbations remain effective at larger scales.}
Gain over base, in accuracy points, when perturbing one component of the first decoder block on Qwen2.5-1.5B and 3B-Instruct for Countdown ($K{=}32$). All components improve over base, with the output and value projections among the strongest at both scales.}
\label{fig:block-qwen-scale}
\end{figure}

\begin{table}[t]
\caption{\textbf{Component-wise gains and selection-reward variability across scales.}
For Qwen2.5-1.5B and 3B-Instruct on Countdown ($K{=}32$), we perturb one component of the first decoder block and report its gain over base and the standard deviation of member rewards on the selection set. Every component produces nonzero reward variability and improves the final vote, with the largest effects generally obtained from the value and output projections.}
\input{tables/tab_component_scale}
\end{table}

\paragraph{Training-reward distributions.} The distribution of member rewards on the $200$ selection examples, measured across all four scales (0.5B, 1.5B, 3B, 7B), shows the same picture. Every component produces a nonzero spread of member rewards, and on OLMo-3-7B-Instruct no component collapses to the base reward across all sampled members. The residual-write components ($\mathtt{o\_proj}$, $\mathtt{down\_proj}$) have a standard deviation of $2$ to $3$ points. At 1.5B, 3B, and 7B, the compute matrices ($\mathtt{q/k/v\_proj}$, $\mathtt{gate/up\_proj}$) have a comparable spread, $1.7$ to $2.5$ points, with the value projection the widest at $2$ to $4$ points; on OLMo, $92$ to $98\%$ of their members improve over base on the selection set. The QKV biases of Qwen spread slightly less, $1.4$ to $2.3$ points at the same scales, and the LayerNorm gains depend on placement, as discussed above. Every component thus induces measurable variation in model behavior, providing nontrivial candidates for selection and voting.

\paragraph{Geometry of the selected perturbations.}
We test the ``dense, connected neighborhood'' picture directly on the OLMo-3-7B Countdown thicket. \emph{Clustering:} the top-$50$ perturbation directions have a mean pairwise cosine of $6{\times}10^{-5}$ (standard deviation $4.6{\times}10^{-3}$), nearly identical to random perturbation pairs ($1{\times}10^{-5}$, standard deviation $4.5{\times}10^{-3}$); the selected directions show no detectable directional clustering by this statistic. \emph{Connectivity:} {averaging two top members with norm preserved and scoring on $200$ held-out test items gives a midpoint accuracy of $69.8\%$ against $69.7\%$ at the endpoints; two random perturbations give $69.3\%$ against $69.4\%$ (base $67.7\%$, $20$ pairs each). The straight path between selected members shows no barrier, so connectivity does not separate them from random perturbations; on held-out items the selected members are barely better than random ones ($69.7\%$ against $69.4\%$).}%
\emph{Generalization:} the best single member improves the selection set by $11$ points but the held-out test set by only $2.4$. Together with the soup collapse (Table~\ref{tab:master}), {the clustering and generalization results} provide little support for interpreting the selected members as a coherent dense weight-space basin. They are instead consistent with selection over dispersed perturbations followed by output-space voting.

\subsection{{Matched Comparison Under the Original Protocol}}
\label{app:upstream}

Table~\ref{tab:upstream} repeats the comparison with the authors' released code and data: their prompts, splits, and scorers, greedy generation of up to $1024$ tokens, $N{=}5000$ and $K{=}50$ for both perturbation families, and a majority vote over $50$ samples at $T{=}0.7$ for TT-MV. Full-weight members draw $\sigma$ from $\{0.001, 0.002, 0.003\}$ as in the original study and logit-bias members from $\{0.25, 0.5, 1, 2, 4, 8\}$; the top $50$ are selected on the $200$ training examples only, so no hyperparameter is chosen on the test set. The two families are statistically indistinguishable on $22$ of the $28$ model--task pairs. Logit PSV is significantly better on Countdown at 1.5B ($+7.2$) and 3B ($+3.6$) and on GSM8K at 1.5B ($+2.1$), and significantly worse on ROCStories at 1.5B ($-1.7$) and 3B ($-0.7$) and on USPTO at 7B ($-2.8$); on OLMo-3-7B-Instruct, MATH-500 ($+2.7$) and OlympiadBench ($+1.9$) are borderline. On USPTO, both families reach exactly the $31.9\%$ of a constant majority-class predictor on every Qwen model. For the Qwen models, most selected logit members use the smallest scale of the grid ($0.25$), so a finer grid could favor logit PSV slightly.

\begin{table}[t]
\caption{{\textbf{Matched comparison under the original protocol} ($N{=}5000$, $K{=}50$, authors' released code). Accuracy (\%) of the greedy base, full-weight PSV, logit PSV, and TT-MV ($T{=}0.7$), and the paired difference between logit and full-weight PSV with its $95\%$ bootstrap confidence interval over test items; bold intervals exclude zero.}}
\input{tables/tab_upstream}
\end{table}

\section{Additional Experiments}
\label{app:experiment}
\subsection{The Full $K$-Sweep Across Methods}
\label{app:kstudy}

We report the complete ensemble-size sweep behind the output-space claims, comparing the three PSV variants head to head at every scale and task. Full-weight PSV perturbs all model weights, logit PSV adds a persistent Gaussian bias to the output logits, and TT-MV performs temperature sampling {at $T{=}0.7$} followed by plurality voting, without reward-based selection. Each is swept over $K \in \{1,2,4,8,16,32,64,128\}$ on four model scales (0.5B, 1.5B, 3B, 7B) and three tasks (Countdown, GSM8K, MATH-500). Figure~\ref{fig:kstudy-acc} shows raw majority-vote accuracy against the greedy base, and Figure~\ref{fig:kstudy-gain} the same runs as gain over base. Across these sweeps, accuracy generally increases with $K$, with larger gains at larger model scales. Logit PSV closely tracks or exceeds full-weight PSV on the reasoning tasks, while TT-MV falls behind primarily in the weak-model regime where reward-based selection is most useful (e.g.\ 0.5B and 1.5B Countdown).

\begin{figure}[p]
\input{figures/fig_kstudy_acc}
\caption{\textbf{Accuracy as a function of ensemble size $K$.}
Plurality-vote accuracy for full-weight PSV, logit PSV, and TT-MV across model scales (rows) and tasks (columns). The dashed line indicates the greedy-base accuracy.%
}
\label{fig:kstudy-acc}
\end{figure}

\begin{figure}[p]
\input{figures/fig_kstudy_gain}
\caption{\textbf{Gain over the greedy base as a function of ensemble size $K$.}
Same runs as Figure~\ref{fig:kstudy-acc}, shown as the accuracy gain $\Delta$ over the greedy base, across model scales (rows) and tasks (columns). The dashed line marks zero gain.
}
\label{fig:kstudy-gain}
\end{figure}

\subsection{Are Weight and Logit Perturbations Complementary?}
\label{app:complementarity}

If weight and logit perturbations provided complementary sources of useful variation, combining them could improve over either intervention alone. We test this directly by composing the two perturbations. For each model and task, we summarize each method by its peak accuracy over the evaluated $K$ values and compare four ensembles: full-weight PSV alone, logit PSV alone, full-weight PSV applied on top of logit PSV (\emph{weight $\circ$ logit}), and, as a control, a second logit perturbation composed with the first using a smaller noise range (\emph{logit $\circ$ logit}). The control tests whether any benefit from combining weight and logit perturbations exceeds that obtained by simply composing two output-space perturbations.

Figure~\ref{fig:compl-bars} reports the corresponding peak gains over base, and Figure~\ref{fig:compl-scatter} compares the combined method with the better single intervention and with the logit-on-logit control. On {nine} of twelve model--task pairs, the combined method matches or trails the better single intervention, and it shows no systematic advantage over the logit-on-logit control. Only {7B MATH-500 improves over both comparisons, by $1.0$ point over the better single intervention and $0.3$ points over the control ($3$ and $1$ of its $300$ test items).} We therefore find little evidence that weight and logit perturbations provide complementary gains, consistent with the view that they largely act through the same output-space mechanism.

\begin{figure}[t]
\input{figures/fig_complementarity_bars}
\caption{\textbf{Accuracy gains from individual and stacked perturbation mechanisms.}
Peak gain over greedy decoding across ensemble sizes $K$, comparing full-weight PSV, logit PSV, their combination, and a same-mechanism control that stacks two logit biases with a smaller noise range. The symbol ``$\circ$'' denotes successive application of two perturbations. Combining weight and logit perturbations rarely improves upon the stronger individual method or the logit-only control, except on {7B MATH-500}.%
}
\label{fig:compl-bars}
\end{figure}

\begin{figure}[t]
\input{figures/fig_complementarity_scatter}
\caption{\textbf{Peak accuracy of stacked perturbations versus individual methods and a logit-only control.}
Full-weight PSV stacked on logit PSV is compared against (a) the stronger individual method and (b) the same-mechanism control, which stacks two logit biases. The dashed diagonal indicates equal accuracy. Most points lie on or below it{. Above it are 3B and 7B MATH-500 ($+1.3$, $+1.0$) and 7B GSM8K ($+0.2$) in (a), and 7B Countdown ($+0.9$), 7B MATH-500 ($+0.3$), and 3B Countdown ($+0.2$) in (b); only 7B MATH-500 is above it in both}.}
\label{fig:compl-scatter}
\end{figure}

%% file: tables/tab_crossscale.tex
\label{tab:crossscale}
\centering
\small
\begin{tabular}{lcccc}
\toprule
task & base & thicket & logit bias & embed bias \\
\midrule
\multicolumn{5}{l}{\textit{Qwen2.5-0.5B-Instruct}}\\
GSM8K & 44.3 & 56.0 & \textbf{56.6} & 55.2 \\
MATH-500 & 29.0 & 39.0 & \textbf{{40.7}} & 35.7 \\
OlympiadBench & 9.5 & 14.3 & \textbf{{16.7}} & 13.7 \\
MBPP & 27.0 & {\textbf{28.8}} & {26.4} & 22.1 \\
ROCStories & 31.8 & 31.9 & {31.9} & 32.1 \\
\midrule
\multicolumn{5}{l}{\textit{Qwen2.5-1.5B-Instruct}}\\
countdown & 12.5 & 37.4 & \textbf{{37.5}} & 39.1 \\
GSM8K & 60.4 & 76.7 & \textbf{{82.0}} & 76.6 \\
MATH-500 & 48.3 & 57.7 & \textbf{{60.0}} & 56.0 \\
OlympiadBench & 21.5 & 28.5 & \textbf{{28.7}} & 28.3 \\
MBPP & 50.0 & {\textbf{52.5}} & {49.1} & 53.3 \\
ROCStories & 46.7 & \textbf{49.0} & {48.2} & 48.9 \\
USPTO{$^{\dagger}$} & 6.3 & 24.9 & {31.3} & 31.9 \\
\midrule
\multicolumn{5}{l}{\textit{Qwen2.5-3B-Instruct}}\\
countdown & 13.7 & {55.2} & \textbf{{57.0}} & 55.8 \\
GSM8K & 81.0 & 87.7 & \textbf{88.7} & 87.1 \\
MATH-500 & 62.7 & {\textbf{72.3}} & {71.7} & 70.3 \\
OlympiadBench & 31.6 & 38.4 & \textbf{{40.5}} & 39.0 \\
MBPP & 59.8 & {\textbf{64.0}} & {63.1} & 63.5 \\
ROCStories & 54.5 & \textbf{56.1} & {55.3} & 55.4 \\
USPTO{$^{\dagger}$} & 7.0 & 19.8 & {15.6} & 15.8 \\
\midrule
\multicolumn{5}{l}{\textit{OLMo-3-7B-Instruct}}\\
countdown & 66.6 & {\textbf{90.8}} & {90.7} & --- \\
GSM8K & 87.5 & 89.8 & \textbf{{91.9}} & --- \\
MATH-500 & 63.3 & 74.7 & \textbf{{81.3}} & --- \\
OlympiadBench & 38.0 & 46.2 & \textbf{{52.5}} & --- \\
MBPP & 64.3 & \textbf{76.4} & {73.2} & --- \\
ROCStories & 65.2 & 65.6 & \textbf{65.8} & --- \\
USPTO{$^{\ddagger}$} & 28.6 & 39.4 & \textbf{{42.7}} & --- \\
\bottomrule
\end{tabular}

%% file: figures/fig_kcurve_qwen.tex
\centering
\begin{tikzpicture}
\begin{groupplot}[
  group style={group size=2 by 3, horizontal sep=1.0cm, vertical sep=1.5cm,
    xticklabels at=edge bottom, yticklabels at=edge left},
  paperaxis, width=0.52\linewidth, height=3.9cm,
  xmode=log, log ticks with fixed point, xtick={1,2,5,10,20,50},
  xmin=0.9, xmax=72,
  xlabel={ensemble size $K$}, ylabel={gain over base (pts)},
  title style={font=\small\bfseries, yshift=-0.5ex},
]
\nextgroupplot[title={Qwen2.5-0.5B, thicket}, ymin=-6, ymax=15,
  legend style={at={(1.07,1.42)}, anchor=south, legend columns=4, font=\footnotesize,
    draw=none, fill=none, /tikz/every even column/.append style={column sep=5pt}}]
\draw[gray,dashed,line width=0.4pt] (axis cs:0.9,0) -- (axis cs:72,0);
\addplot[black, forget plot]  coordinates {(1,-4.10)(2,-3.87)(5,5.45)(10,8.64)(20,10.16)(50,11.75)};
\addplot[csoup, forget plot]  coordinates {(1,-3.67)(2,-3.33)(5,2.67)(10,6.67)(20,9.33)(50,10.00)};
\addplot[cextra, forget plot] coordinates {(1,-2.11)(2,-2.11)(5,-0.84)(10,1.48)(20,3.59)(50,4.86)};
\addplot[cttmv, forget plot]  coordinates {(1,1.84)(2,1.97)(5,2.62)(10,2.88)(20,3.14)(50,1.84)};
\addplot[gray, forget plot]   coordinates {(1,0.24)(2,0.24)(5,0.20)(10,0.27)(20,0.03)(50,0.04)};
\addlegendimage{cthicket}\addlegendentry{countdown}
\addlegendimage{black}\addlegendentry{GSM8K}
\addlegendimage{csoup}\addlegendentry{MATH-500}
\addlegendimage{cextra}\addlegendentry{OlympiadBench}
\addlegendimage{cttmv}\addlegendentry{MBPP}
\addlegendimage{gray}\addlegendentry{ROCStories}
\addlegendimage{clogit}\addlegendentry{USPTO}
\nextgroupplot[title={Qwen2.5-0.5B, logit bias}, ymin=-6, ymax=15]
\draw[gray,dashed,line width=0.4pt] (axis cs:0.9,0) -- (axis cs:72,0);
\addplot[black]  coordinates {(1,-2.12)(2,-1.97)(5,5.84)(10,8.64)(20,10.77)(50,12.28)};
\addplot[csoup]  coordinates {(1,-2.33)(2,-2.33)(5,5.67)(10,8.33)(20,10.33)(50,11.67)};
\addplot[cextra] coordinates {(1,-0.84)(2,-0.63)(5,1.69)(10,4.85)(20,4.64)(50,7.17)};
\addplot[cttmv]  coordinates {(1,2.23)(2,2.49)(5,2.23)(10,2.88)(20,1.05)(50,-0.52)};
\addplot[gray]   coordinates {(1,0.56)(2,0.56)(5,0.05)(10,0.58)(20,-0.40)(50,0.06)};
\nextgroupplot[title={Qwen2.5-1.5B, thicket}, ymin=-9, ymax=30]
\draw[gray,dashed,line width=0.4pt] (axis cs:0.9,0) -- (axis cs:72,0);
\addplot[cthicket] coordinates {(1,1.05)(2,7.15)(5,16.40)(10,20.85)(20,23.95)(50,24.90)};
\addplot[black]    coordinates {(1,8.95)(2,8.95)(5,12.89)(10,15.77)(20,16.23)(50,16.30)};
\addplot[csoup]    coordinates {(1,-0.33)(2,-0.33)(5,4.34)(10,6.34)(20,8.67)(50,9.34)};
\addplot[cextra]   coordinates {(1,-3.17)(2,-2.95)(5,3.37)(10,6.12)(20,6.96)(50,6.96)};
\addplot[cttmv]    coordinates {(1,-0.26)(2,-0.26)(5,0.65)(10,1.05)(20,3.14)(50,2.49)};
\addplot[gray]     coordinates {(1,2.50)(2,2.50)(5,2.45)(10,2.19)(20,2.32)(50,2.28)};
\addplot[clogit]   coordinates {(1,20.38)(2,20.38)(5,25.18)(10,23.88)(20,23.98)(50,18.59)};
\nextgroupplot[title={Qwen2.5-1.5B, logit bias}, ymin=-9, ymax=30]
\draw[gray,dashed,line width=0.4pt] (axis cs:0.9,0) -- (axis cs:72,0);
\addplot[cthicket] coordinates {(1,-1.35)(2,3.60)(5,10.40)(10,16.40)(20,21.10)(50,25.00)};
\addplot[black]    coordinates {(1,6.44)(2,6.97)(5,15.39)(10,17.97)(20,20.92)(50,21.53)};
\addplot[csoup]    coordinates {(1,-6.66)(2,-6.33)(5,2.67)(10,7.34)(20,10.00)(50,11.67)};
\addplot[cextra]   coordinates {(1,-5.70)(2,-4.85)(5,-0.84)(10,3.59)(20,6.75)(50,7.17)};
\addplot[cttmv]    coordinates {(1,-6.54)(2,-5.89)(5,-5.24)(10,-4.84)(20,-3.27)(50,-0.92)};
\addplot[gray]     coordinates {(1,1.80)(2,1.87)(5,1.92)(10,1.52)(20,1.57)(50,1.52)};
\addplot[clogit]   coordinates {(1,25.37)(2,25.47)(5,25.67)(10,25.47)(20,25.57)(50,24.98)};
\nextgroupplot[title={Qwen2.5-3B, thicket}, ymin=-8, ymax=45]
\draw[gray,dashed,line width=0.4pt] (axis cs:0.9,0) -- (axis cs:72,0);
\addplot[cthicket] coordinates {(1,5.45)(2,16.25)(5,27.25)(10,33.20)(20,38.70)(50,41.50)};
\addplot[black]    coordinates {(1,-0.23)(2,-0.23)(5,4.70)(10,5.76)(20,6.44)(50,6.75)};
\addplot[csoup]    coordinates {(1,-3.67)(2,-3.67)(5,0.33)(10,5.66)(20,8.66)(50,9.66)};
\addplot[cextra]   coordinates {(1,-4.65)(2,-4.43)(5,1.68)(10,5.27)(20,6.54)(50,6.75)};
\addplot[cttmv]    coordinates {(1,0.26)(2,0.39)(5,1.44)(10,2.61)(20,3.66)(50,4.19)};
\addplot[gray]     coordinates {(1,1.53)(2,1.89)(5,1.86)(10,1.85)(20,1.80)(50,1.62)};
\addplot[clogit]   coordinates {(1,19.78)(2,19.78)(5,20.18)(10,19.48)(20,16.49)(50,12.79)};
\nextgroupplot[title={Qwen2.5-3B, logit bias}, ymin=-8, ymax=45]
\draw[gray,dashed,line width=0.4pt] (axis cs:0.9,0) -- (axis cs:72,0);
\addplot[cthicket] coordinates {(1,15.15)(2,21.85)(5,31.35)(10,36.10)(20,39.60)(50,43.30)};
\addplot[black]    coordinates {(1,-0.83)(2,-0.83)(5,5.91)(10,6.97)(20,7.66)(50,7.73)};
\addplot[csoup]    coordinates {(1,-6.67)(2,-6.67)(5,2.67)(10,7.00)(20,9.00)(50,9.00)};
\addplot[cextra]   coordinates {(1,-5.49)(2,-4.64)(5,1.48)(10,4.01)(20,7.38)(50,8.86)};
\addplot[cttmv]    coordinates {(1,-1.18)(2,-1.05)(5,-0.13)(10,0.79)(20,1.70)(50,3.27)};
\addplot[gray]     coordinates {(1,0.40)(2,0.46)(5,0.63)(10,0.67)(20,0.79)(50,0.80)};
\addplot[clogit]   coordinates {(1,15.99)(2,15.99)(5,12.39)(10,11.19)(20,9.49)(50,8.59)};
\end{groupplot}
\end{tikzpicture}

%% file: figures/fig_block_qwen.tex
\centering
\ifdefined\qwendiagram\else\input{figures/qwen_block_diagram}\fi
\resizebox{\linewidth}{!}{\qwendiagram{+1.4}{+1.4}{+1.4}{+7.6}{+5.5}{+1.7}{+4.2}{+7.1}{+5.5}}\\[4pt]
{\scriptsize
\mbox{\qwenkeywrite\ writes into the stream\ (\qW{$+5.5$})}\qquad
\mbox{\qwenkeyln\ LayerNorm gain, \emph{pre}-norm\ (\qQ{$+1$} to \qQ{$+2$})}\qquad
\mbox{\qwenkeycmp\ compute matrices\ (\qC{$+1.4$} to \qC{$+7.6$})}}

%% file: figures/fig_block_qwen_scale.tex
\centering
\ifdefined\qwendiagram\else\input{figures/qwen_block_diagram}\fi
\textbf{\footnotesize (a) Qwen2.5-1.5B-Instruct \ (countdown, base $12.5$, full block $+24.9$)}\\[2pt]
\resizebox{\linewidth}{!}{\qwendiagram{+12.3}{+14.3}{+15.9}{+26.3}{+26.9}{+11.7}{+16.4}{+18.4}{+21.7}}\\[8pt]
\textbf{\footnotesize (b) Qwen2.5-3B-Instruct \ (countdown, base $13.7$, full block $+40.3$)}\\[2pt]
\resizebox{\linewidth}{!}{\qwendiagram{+29.8}{+30.9}{+32.1}{+37.4}{+39.9}{+29.5}{+33.4}{+35.1}{+37.5}}\\[4pt]
{\scriptsize
\qwenkeywrite\ writes into the stream\qquad
\qwenkeyln\ LayerNorm gain (pre-norm)\qquad
\qwenkeycmp\ compute matrices}

%% file: tables/tab_component_scale.tex
\label{tab:component-scale}
\centering
\small
\begin{tabular}{lcccc}
\toprule
 & \multicolumn{2}{c}{1.5B} & \multicolumn{2}{c}{3B} \\
\cmidrule(lr){2-3}\cmidrule(lr){4-5}
single weight perturbed & gain & reward std & gain & reward std \\
\midrule
\multicolumn{5}{l}{\textit{residual writes}}\\
\texttt{self\_attn.o\_proj} & $+26.9$ & $2.6$ & $+39.9$ & $3.0$ \\
\texttt{mlp.down\_proj} & $+21.7$ & $2.2$ & $+37.5$ & $2.4$ \\
\multicolumn{5}{l}{\textit{LayerNorm gains (pre-norm)}}\\
\texttt{input\_layernorm} & $+12.3$ & $1.4$ & $+29.8$ & $2.2$ \\
\texttt{post\_attention\_layernorm} & $+11.7$ & $1.4$ & $+29.5$ & $2.2$ \\
\multicolumn{5}{l}{\textit{compute matrices}}\\
\texttt{self\_attn.q\_proj} & $+14.3$ & $1.7$ & $+30.9$ & $2.3$ \\
\texttt{self\_attn.k\_proj} & $+15.9$ & $1.7$ & $+32.1$ & $2.3$ \\
\texttt{self\_attn.v\_proj} & $+26.3$ & $3.6$ & $+37.4$ & $4.2$ \\
\texttt{mlp.gate\_proj} & $+16.4$ & $1.8$ & $+33.4$ & $2.4$ \\
\texttt{mlp.up\_proj} & $+18.4$ & $2.0$ & $+35.1$ & $2.5$ \\
\multicolumn{5}{l}{\textit{biases}}\\
\texttt{self\_attn.q\_proj.bias} & $+11.0$ & $1.4$ & $+28.1$ & $2.2$ \\
\texttt{self\_attn.k\_proj.bias} & $+10.9$ & $1.4$ & $+27.5$ & $2.2$ \\
\texttt{self\_attn.v\_proj.bias} & $+12.4$ & $1.5$ & $+30.7$ & $2.3$ \\
\bottomrule
\end{tabular}

%% file: tables/tab_upstream.tex
\label{tab:upstream}
\centering
\footnotesize
{
\begin{tabular}{lccccc}
\toprule
task & base & full-weight PSV & logit PSV & TT-MV & logit $-$ full-weight [95\% CI] \\
\midrule
\multicolumn{6}{l}{\textit{Qwen2.5-0.5B-Instruct}}\\
countdown & 0.3 & 12.6 & 12.8 & 7.2 & $+$0.2 [$-$0.8, $+$1.1] \\
GSM8K & 44.3 & 55.3 & 54.4 & 57.7 & $-$1.0 [$-$2.7, $+$0.7] \\
MATH-500 & 29.0 & 39.3 & 39.3 & 43.0 & $+$0.0 [$-$3.7, $+$4.0] \\
OlympiadBench & 7.2 & 15.2 & 14.8 & 14.8 & $-$0.4 [$-$2.7, $+$1.7] \\
MBPP & 27.0 & 27.2 & 28.5 & 11.8 & $+$1.3 [$-$1.3, $+$3.8] \\
ROCStories & 32.0 & 32.1 & 31.9 & 32.0 & $-$0.2 [$-$0.6, $+$0.2] \\
USPTO & 31.9 & 31.9 & 31.9 & 31.9 & $+$0.0 [$+$0.0, $+$0.0] \\
\midrule
\multicolumn{6}{l}{\textit{Qwen2.5-1.5B-Instruct}}\\
countdown & 12.5 & 33.2 & 40.4 & 24.9 & \textbf{$+$7.2 [$+$5.6, $+$8.8]} \\
GSM8K & 60.4 & 77.9 & 80.0 & 79.2 & \textbf{$+$2.1 [$+$1.0, $+$3.4]} \\
MATH-500 & 47.3 & 57.3 & 57.0 & 59.3 & $-$0.3 [$-$3.0, $+$2.3] \\
OlympiadBench & 20.3 & 28.5 & 29.1 & 30.4 & $+$0.6 [$-$1.1, $+$2.3] \\
MBPP & 50.0 & 52.1 & 51.3 & 46.2 & $-$0.8 [$-$2.6, $+$1.2] \\
ROCStories & 46.7 & 49.7 & 48.0 & 46.9 & \textbf{$-$1.7 [$-$2.0, $-$1.4]} \\
USPTO & 6.3 & 31.9 & 31.9 & 2.5 & $+$0.0 [$+$0.0, $+$0.0] \\
\midrule
\multicolumn{6}{l}{\textit{Qwen2.5-3B-Instruct}}\\
countdown & 13.7 & 53.5 & 57.1 & 56.1 & \textbf{$+$3.6 [$+$2.1, $+$5.0]} \\
GSM8K & 81.0 & 87.9 & 87.6 & 88.5 & $-$0.4 [$-$1.3, $+$0.6] \\
MATH-500 & 58.7 & 68.7 & 69.7 & 70.7 & $+$1.0 [$-$1.0, $+$3.3] \\
OlympiadBench & 32.5 & 39.9 & 39.2 & 40.9 & $-$0.6 [$-$2.7, $+$1.5] \\
MBPP & 59.8 & 63.0 & 63.6 & 62.4 & $+$0.7 [$-$1.7, $+$3.1] \\
ROCStories & 54.7 & 56.5 & 55.8 & 55.0 & \textbf{$-$0.7 [$-$0.9, $-$0.4]} \\
USPTO & 7.0 & 31.9 & 31.9 & 7.4 & $+$0.0 [$-$0.3, $+$0.3] \\
\midrule
\multicolumn{6}{l}{\textit{OLMo-3-7B-Instruct}}\\
countdown & 66.6 & 91.8 & 91.6 & 92.0 & $-$0.1 [$-$0.9, $+$0.7] \\
GSM8K & 87.5 & 89.6 & 90.0 & 89.0 & $+$0.4 [$-$0.4, $+$1.1] \\
MATH-500 & 51.0 & 70.0 & 72.7 & 67.3 & $+$2.7 [$+$0.0, $+$5.3] \\
OlympiadBench & 27.0 & 38.0 & 39.9 & 38.2 & $+$1.9 [$+$0.0, $+$3.8] \\
MBPP & 62.8 & 73.2 & 72.6 & 73.6 & $-$0.5 [$-$2.7, $+$1.6] \\
ROCStories & 64.0 & 64.7 & 64.7 & 64.2 & $+$0.0 [$-$0.2, $+$0.3] \\
USPTO & 28.6 & 42.6 & 39.8 & 36.3 & \textbf{$-$2.8 [$-$4.6, $-$1.0]} \\
\bottomrule
\end{tabular}
}

%% file: figures/fig_kstudy_acc.tex
\centering
\begin{tikzpicture}
\begin{groupplot}[
  group style={group size=3 by 4, horizontal sep=1.15cm, vertical sep=1.25cm,
    xticklabels at=edge bottom},
  paperaxis, width=0.37\linewidth, height=3.55cm,
  xmode=log, log basis x=2, log ticks with fixed point,
  xtick={1,4,16,64}, xmin=0.85, xmax=150,
  xlabel={ensemble size $K$},
  title style={font=\small\bfseries, yshift=-0.5ex},
]
\nextgroupplot[title={0.5B, Countdown}, ymin=-5, ymax=15, ylabel={accuracy (\%)}, legend style={at={(1.72,1.30)}, anchor=south, legend columns=-1, draw=none, fill=none, /tikz/every even column/.append style={column sep=9pt}}]
\addplot[cbase,dashed,line width=0.7pt,forget plot] coordinates {(0.85,0.30)(150,0.30)};
\addplot[cthicket, mark=*, mark size=1.4pt, forget plot] coordinates {(1,3.85)(2,7.50)(4,8.85)(8,10.00)(16,10.85)(32,12.15)(64,12.50)(128,12.75)};
\addplot[clogit, mark=square*, mark size=1.3pt, forget plot] coordinates {(1,6.05)(2,8.55)(4,10.80)(8,11.65)(16,11.65)(32,12.30)(64,12.90)(128,12.60)};
\addplot[cttmv, mark=triangle*, mark size=1.6pt, forget plot] coordinates {(1,1.25)(2,1.35)(4,1.35)(8,2.80)(16,4.75)(32,6.90)(64,8.25)(128,10.35)};
\addlegendimage{cthicket, mark=*, mark size=1.4pt}\addlegendentry{full-weight thicket}
\addlegendimage{clogit, mark=square*, mark size=1.3pt}\addlegendentry{logit bias}
\addlegendimage{cttmv, mark=triangle*, mark size=1.6pt}\addlegendentry{TT-MV}
\addlegendimage{cbase, dashed, line width=0.7pt}\addlegendentry{greedy base}
\nextgroupplot[title={0.5B, GSM8K}, ymin=35, ymax=60]
\addplot[cbase,dashed,line width=0.7pt,forget plot] coordinates {(0.85,44.28)(150,44.28)};
\addplot[cthicket, mark=*, mark size=1.4pt, forget plot] coordinates {(1,40.18)(2,40.41)(4,47.31)(8,51.33)(16,53.15)(32,55.34)(64,55.72)(128,55.95)};
\addplot[clogit, mark=square*, mark size=1.3pt, forget plot] coordinates {(1,42.15)(2,42.30)(4,48.90)(8,52.31)(16,55.88)(32,56.41)(64,56.71)(128,57.32)};
\addplot[cttmv, mark=triangle*, mark size=1.6pt, forget plot] coordinates {(1,39.35)(2,39.42)(4,46.10)(8,52.54)(16,55.50)(32,56.79)(64,57.39)(128,57.85)};
\nextgroupplot[title={0.5B, MATH-500}, ymin=20, ymax=45]
\addplot[cbase,dashed,line width=0.7pt,forget plot] coordinates {(0.85,29.00)(150,29.00)};
\addplot[cthicket, mark=*, mark size=1.4pt, forget plot] coordinates {(1,25.33)(2,25.67)(4,30.00)(8,35.00)(16,37.33)(32,37.67)(64,39.00)(128,39.33)};
\addplot[clogit, mark=square*, mark size=1.3pt, forget plot] coordinates {(1,26.67)(2,26.67)(4,33.67)(8,36.67)(16,39.00)(32,40.00)(64,40.67)(128,42.00)};
\addplot[cttmv, mark=triangle*, mark size=1.6pt, forget plot] coordinates {(1,26.00)(2,26.00)(4,32.00)(8,37.67)(16,41.33)(32,40.33)(64,41.33)(128,41.33)};
\nextgroupplot[title={1.5B, Countdown}, ymin=5, ymax=45, ylabel={accuracy (\%)}]
\addplot[cbase,dashed,line width=0.7pt,forget plot] coordinates {(0.85,12.50)(150,12.50)};
\addplot[cthicket, mark=*, mark size=1.4pt, forget plot] coordinates {(1,13.55)(2,19.65)(4,26.60)(8,31.90)(16,35.65)(32,36.05)(64,38.15)(128,38.65)};
\addplot[clogit, mark=square*, mark size=1.3pt, forget plot] coordinates {(1,18.10)(2,22.30)(4,27.80)(8,31.40)(16,34.45)(32,38.50)(64,39.45)(128,40.45)};
\addplot[cttmv, mark=triangle*, mark size=1.6pt, forget plot] coordinates {(1,8.00)(2,11.00)(4,14.25)(8,18.50)(16,19.65)(32,24.20)(64,25.80)(128,28.60)};
\nextgroupplot[title={1.5B, GSM8K}, ymin=55, ymax=85]
\addplot[cbase,dashed,line width=0.7pt,forget plot] coordinates {(0.85,60.42)(150,60.42)};
\addplot[cthicket, mark=*, mark size=1.4pt, forget plot] coordinates {(1,69.37)(2,69.37)(4,72.48)(8,75.28)(16,76.35)(32,76.65)(64,77.10)(128,76.65)};
\addplot[clogit, mark=square*, mark size=1.3pt, forget plot] coordinates {(1,66.87)(2,67.40)(4,74.07)(8,77.48)(16,80.97)(32,81.80)(64,82.49)(128,82.34)};
\addplot[cttmv, mark=triangle*, mark size=1.6pt, forget plot] coordinates {(1,57.54)(2,57.62)(4,67.93)(8,74.37)(16,75.74)(32,78.32)(64,79.76)(128,80.74)};
\nextgroupplot[title={1.5B, MATH-500}, ymin=35, ymax=65]
\addplot[cbase,dashed,line width=0.7pt,forget plot] coordinates {(0.85,48.33)(150,48.33)};
\addplot[cthicket, mark=*, mark size=1.4pt, forget plot] coordinates {(1,48.00)(2,48.00)(4,53.33)(8,53.33)(16,56.33)(32,56.67)(64,59.33)(128,56.67)};
\addplot[clogit, mark=square*, mark size=1.3pt, forget plot] coordinates {(1,41.67)(2,42.00)(4,48.33)(8,54.33)(16,57.33)(32,61.33)(64,61.67)(128,61.33)};
\addplot[cttmv, mark=triangle*, mark size=1.6pt, forget plot] coordinates {(1,39.00)(2,39.33)(4,50.33)(8,52.67)(16,55.67)(32,58.33)(64,58.33)(128,59.00)};
\nextgroupplot[title={3B, Countdown}, ymin=10, ymax=65, ylabel={accuracy (\%)}]
\addplot[cbase,dashed,line width=0.7pt,forget plot] coordinates {(0.85,13.65)(150,13.65)};
\addplot[cthicket, mark=*, mark size=1.4pt, forget plot] coordinates {(1,22.35)(2,31.80)(4,40.90)(8,47.75)(16,52.05)(32,55.55)(64,56.80)(128,57.65)};
\addplot[clogit, mark=square*, mark size=1.3pt, forget plot] coordinates {(1,21.50)(2,33.05)(4,38.15)(8,45.70)(16,51.65)(32,53.60)(64,56.45)(128,57.90)};
\addplot[cttmv, mark=triangle*, mark size=1.6pt, forget plot] coordinates {(1,14.50)(2,24.50)(4,34.85)(8,42.10)(16,50.30)(32,54.10)(64,57.20)(128,57.45)};
\nextgroupplot[title={3B, GSM8K}, ymin=75, ymax=90]
\addplot[cbase,dashed,line width=0.7pt,forget plot] coordinates {(0.85,80.97)(150,80.97)};
\addplot[cthicket, mark=*, mark size=1.4pt, forget plot] coordinates {(1,80.74)(2,80.74)(4,84.76)(8,86.66)(16,87.19)(32,87.34)(64,87.41)(128,87.49)};
\addplot[clogit, mark=square*, mark size=1.3pt, forget plot] coordinates {(1,80.14)(2,80.14)(4,85.37)(8,87.72)(16,88.48)(32,88.86)(64,88.86)(128,89.31)};
\addplot[cttmv, mark=triangle*, mark size=1.6pt, forget plot] coordinates {(1,78.39)(2,78.47)(4,84.00)(8,85.52)(16,87.19)(32,88.48)(64,88.63)(128,88.70)};
\nextgroupplot[title={3B, MATH-500}, ymin=55, ymax=80]
\addplot[cbase,dashed,line width=0.7pt,forget plot] coordinates {(0.85,62.67)(150,62.67)};
\addplot[cthicket, mark=*, mark size=1.4pt, forget plot] coordinates {(1,59.00)(2,59.00)(4,61.33)(8,64.67)(16,69.67)(32,72.00)(64,71.67)(128,72.00)};
\addplot[clogit, mark=square*, mark size=1.3pt, forget plot] coordinates {(1,56.00)(2,56.00)(4,62.67)(8,68.33)(16,72.00)(32,72.67)(64,72.67)(128,72.67)};
\addplot[cttmv, mark=triangle*, mark size=1.6pt, forget plot] coordinates {(1,62.00)(2,62.00)(4,65.00)(8,71.33)(16,73.67)(32,73.00)(64,73.33)(128,73.00)};
\nextgroupplot[title={7B, Countdown}, ymin=55, ymax=100, ylabel={accuracy (\%)}]
\addplot[cbase,dashed,line width=0.7pt,forget plot] coordinates {(0.85,66.60)(150,66.60)};
\addplot[cthicket, mark=*, mark size=1.4pt, forget plot] coordinates {(1,68.15)(2,74.50)(4,79.95)(8,84.00)(16,87.80)(32,90.35)(64,92.35)(128,93.95)};
\addplot[clogit, mark=square*, mark size=1.3pt, forget plot] coordinates {(1,60.35)(2,69.00)(4,76.00)(8,80.70)(16,84.40)(32,88.85)(64,91.50)(128,93.85)};
\addplot[cttmv, mark=triangle*, mark size=1.6pt, forget plot] coordinates {(1,66.45)(2,75.15)(4,80.55)(8,84.40)(16,88.20)(32,90.85)(64,92.75)(128,94.25)};
\nextgroupplot[title={7B, GSM8K}, ymin=85, ymax=95]
\addplot[cbase,dashed,line width=0.7pt,forget plot] coordinates {(0.85,87.49)(150,87.49)};
\addplot[cthicket, mark=*, mark size=1.4pt, forget plot] coordinates {(1,88.32)(2,88.93)(4,89.76)(8,89.54)(16,89.46)(32,89.69)(64,89.92)(128,89.76)};
\addplot[clogit, mark=square*, mark size=1.3pt, forget plot] coordinates {(1,86.28)(2,86.66)(4,89.99)(8,90.98)(16,91.51)(32,91.96)(64,91.96)(128,91.89)};
\addplot[cttmv, mark=triangle*, mark size=1.6pt, forget plot] coordinates {(1,86.58)(2,87.04)(4,88.02)(8,89.23)(16,89.31)(32,89.39)(64,89.16)(128,89.46)};
\nextgroupplot[title={7B, MATH-500}, ymin=50, ymax=85]
\addplot[cbase,dashed,line width=0.7pt,forget plot] coordinates {(0.85,63.33)(150,63.33)};
\addplot[cthicket, mark=*, mark size=1.4pt, forget plot] coordinates {(1,64.00)(2,64.33)(4,72.00)(8,76.33)(16,78.33)(32,81.00)(64,80.67)(128,80.33)};
\addplot[clogit, mark=square*, mark size=1.3pt, forget plot] coordinates {(1,72.33)(2,73.00)(4,77.00)(8,78.00)(16,80.00)(32,81.33)(64,81.67)(128,82.00)};
\addplot[cttmv, mark=triangle*, mark size=1.6pt, forget plot] coordinates {(1,55.67)(2,58.00)(4,67.00)(8,72.67)(16,75.00)(32,77.00)(64,77.67)(128,78.00)};
\end{groupplot}
\end{tikzpicture}

%% file: figures/fig_kstudy_gain.tex
\centering
\begin{tikzpicture}
\begin{groupplot}[
  group style={group size=3 by 4, horizontal sep=1.15cm, vertical sep=1.25cm,
    xticklabels at=edge bottom},
  paperaxis, width=0.37\linewidth, height=3.55cm,
  xmode=log, log basis x=2, log ticks with fixed point,
  xtick={1,4,16,64}, xmin=0.85, xmax=150,
  xlabel={ensemble size $K$},
  title style={font=\small\bfseries, yshift=-0.5ex},
]
\nextgroupplot[title={0.5B, Countdown}, ymin=-5, ymax=15, ylabel={gain over base (pts)}, legend style={at={(1.72,1.30)}, anchor=south, legend columns=-1, draw=none, fill=none, /tikz/every even column/.append style={column sep=9pt}}]
\draw[cbase,dashed,line width=0.5pt] (axis cs:0.85,0) -- (axis cs:150,0);
\addplot[cthicket, mark=*, mark size=1.4pt, forget plot] coordinates {(1,3.55)(2,7.20)(4,8.55)(8,9.70)(16,10.55)(32,11.85)(64,12.20)(128,12.45)};
\addplot[clogit, mark=square*, mark size=1.3pt, forget plot] coordinates {(1,5.75)(2,8.25)(4,10.50)(8,11.35)(16,11.35)(32,12.00)(64,12.60)(128,12.30)};
\addplot[cttmv, mark=triangle*, mark size=1.6pt, forget plot] coordinates {(1,0.95)(2,1.05)(4,1.05)(8,2.50)(16,4.45)(32,6.60)(64,7.95)(128,10.05)};
\addlegendimage{cthicket, mark=*, mark size=1.4pt}\addlegendentry{full-weight thicket}
\addlegendimage{clogit, mark=square*, mark size=1.3pt}\addlegendentry{logit bias}
\addlegendimage{cttmv, mark=triangle*, mark size=1.6pt}\addlegendentry{TT-MV}
\nextgroupplot[title={0.5B, GSM8K}, ymin=-10, ymax=20]
\draw[cbase,dashed,line width=0.5pt] (axis cs:0.85,0) -- (axis cs:150,0);
\addplot[cthicket, mark=*, mark size=1.4pt, forget plot] coordinates {(1,-4.10)(2,-3.87)(4,3.03)(8,7.05)(16,8.87)(32,11.06)(64,11.44)(128,11.67)};
\addplot[clogit, mark=square*, mark size=1.3pt, forget plot] coordinates {(1,-2.13)(2,-1.98)(4,4.62)(8,8.03)(16,11.60)(32,12.13)(64,12.43)(128,13.04)};
\addplot[cttmv, mark=triangle*, mark size=1.6pt, forget plot] coordinates {(1,-4.93)(2,-4.86)(4,1.82)(8,8.26)(16,11.22)(32,12.51)(64,13.11)(128,13.57)};
\nextgroupplot[title={0.5B, MATH-500}, ymin=-10, ymax=20]
\draw[cbase,dashed,line width=0.5pt] (axis cs:0.85,0) -- (axis cs:150,0);
\addplot[cthicket, mark=*, mark size=1.4pt, forget plot] coordinates {(1,-3.67)(2,-3.33)(4,1.00)(8,6.00)(16,8.33)(32,8.67)(64,10.00)(128,10.33)};
\addplot[clogit, mark=square*, mark size=1.3pt, forget plot] coordinates {(1,-2.33)(2,-2.33)(4,4.67)(8,7.67)(16,10.00)(32,11.00)(64,11.67)(128,13.00)};
\addplot[cttmv, mark=triangle*, mark size=1.6pt, forget plot] coordinates {(1,-3.00)(2,-3.00)(4,3.00)(8,8.67)(16,12.33)(32,11.33)(64,12.33)(128,12.33)};
\nextgroupplot[title={1.5B, Countdown}, ymin=-5, ymax=35, ylabel={gain over base (pts)}]
\draw[cbase,dashed,line width=0.5pt] (axis cs:0.85,0) -- (axis cs:150,0);
\addplot[cthicket, mark=*, mark size=1.4pt, forget plot] coordinates {(1,1.05)(2,7.15)(4,14.10)(8,19.40)(16,23.15)(32,23.55)(64,25.65)(128,26.15)};
\addplot[clogit, mark=square*, mark size=1.3pt, forget plot] coordinates {(1,5.60)(2,9.80)(4,15.30)(8,18.90)(16,21.95)(32,26.00)(64,26.95)(128,27.95)};
\addplot[cttmv, mark=triangle*, mark size=1.6pt, forget plot] coordinates {(1,-4.50)(2,-1.50)(4,1.75)(8,6.00)(16,7.15)(32,11.70)(64,13.30)(128,16.10)};
\nextgroupplot[title={1.5B, GSM8K}, ymin=-5, ymax=25]
\draw[cbase,dashed,line width=0.5pt] (axis cs:0.85,0) -- (axis cs:150,0);
\addplot[cthicket, mark=*, mark size=1.4pt, forget plot] coordinates {(1,8.95)(2,8.95)(4,12.06)(8,14.86)(16,15.93)(32,16.23)(64,16.68)(128,16.23)};
\addplot[clogit, mark=square*, mark size=1.3pt, forget plot] coordinates {(1,6.45)(2,6.98)(4,13.65)(8,17.06)(16,20.55)(32,21.38)(64,22.07)(128,21.92)};
\addplot[cttmv, mark=triangle*, mark size=1.6pt, forget plot] coordinates {(1,-2.88)(2,-2.80)(4,7.51)(8,13.95)(16,15.32)(32,17.90)(64,19.34)(128,20.32)};
\nextgroupplot[title={1.5B, MATH-500}, ymin=-15, ymax=15]
\draw[cbase,dashed,line width=0.5pt] (axis cs:0.85,0) -- (axis cs:150,0);
\addplot[cthicket, mark=*, mark size=1.4pt, forget plot] coordinates {(1,-0.33)(2,-0.33)(4,5.00)(8,5.00)(16,8.00)(32,8.34)(64,11.00)(128,8.34)};
\addplot[clogit, mark=square*, mark size=1.3pt, forget plot] coordinates {(1,-6.66)(2,-6.33)(4,0.00)(8,6.00)(16,9.00)(32,13.00)(64,13.34)(128,13.00)};
\addplot[cttmv, mark=triangle*, mark size=1.6pt, forget plot] coordinates {(1,-9.33)(2,-9.00)(4,2.00)(8,4.34)(16,7.34)(32,10.00)(64,10.00)(128,10.67)};
\nextgroupplot[title={3B, Countdown}, ymin=-5, ymax=50, ylabel={gain over base (pts)}]
\draw[cbase,dashed,line width=0.5pt] (axis cs:0.85,0) -- (axis cs:150,0);
\addplot[cthicket, mark=*, mark size=1.4pt, forget plot] coordinates {(1,8.70)(2,18.15)(4,27.25)(8,34.10)(16,38.40)(32,41.90)(64,43.15)(128,44.00)};
\addplot[clogit, mark=square*, mark size=1.3pt, forget plot] coordinates {(1,7.85)(2,19.40)(4,24.50)(8,32.05)(16,38.00)(32,39.95)(64,42.80)(128,44.25)};
\addplot[cttmv, mark=triangle*, mark size=1.6pt, forget plot] coordinates {(1,0.85)(2,10.85)(4,21.20)(8,28.45)(16,36.65)(32,40.45)(64,43.55)(128,43.80)};
\nextgroupplot[title={3B, GSM8K}, ymin=-5, ymax=10]
\draw[cbase,dashed,line width=0.5pt] (axis cs:0.85,0) -- (axis cs:150,0);
\addplot[cthicket, mark=*, mark size=1.4pt, forget plot] coordinates {(1,-0.23)(2,-0.23)(4,3.79)(8,5.69)(16,6.22)(32,6.37)(64,6.44)(128,6.52)};
\addplot[clogit, mark=square*, mark size=1.3pt, forget plot] coordinates {(1,-0.83)(2,-0.83)(4,4.40)(8,6.75)(16,7.51)(32,7.89)(64,7.89)(128,8.34)};
\addplot[cttmv, mark=triangle*, mark size=1.6pt, forget plot] coordinates {(1,-2.58)(2,-2.50)(4,3.03)(8,4.55)(16,6.22)(32,7.51)(64,7.66)(128,7.73)};
\nextgroupplot[title={3B, MATH-500}, ymin=-8, ymax=15]
\draw[cbase,dashed,line width=0.5pt] (axis cs:0.85,0) -- (axis cs:150,0);
\addplot[cthicket, mark=*, mark size=1.4pt, forget plot] coordinates {(1,-3.67)(2,-3.67)(4,-1.34)(8,2.00)(16,7.00)(32,9.33)(64,9.00)(128,9.33)};
\addplot[clogit, mark=square*, mark size=1.3pt, forget plot] coordinates {(1,-6.67)(2,-6.67)(4,0.00)(8,5.66)(16,9.33)(32,10.00)(64,10.00)(128,10.00)};
\addplot[cttmv, mark=triangle*, mark size=1.6pt, forget plot] coordinates {(1,-0.67)(2,-0.67)(4,2.33)(8,8.66)(16,11.00)(32,10.33)(64,10.66)(128,10.33)};
\nextgroupplot[title={7B, Countdown}, ymin=-10, ymax=35, ylabel={gain over base (pts)}]
\draw[cbase,dashed,line width=0.5pt] (axis cs:0.85,0) -- (axis cs:150,0);
\addplot[cthicket, mark=*, mark size=1.4pt, forget plot] coordinates {(1,1.55)(2,7.90)(4,13.35)(8,17.40)(16,21.20)(32,23.75)(64,25.75)(128,27.35)};
\addplot[clogit, mark=square*, mark size=1.3pt, forget plot] coordinates {(1,-6.25)(2,2.40)(4,9.40)(8,14.10)(16,17.80)(32,22.25)(64,24.90)(128,27.25)};
\addplot[cttmv, mark=triangle*, mark size=1.6pt, forget plot] coordinates {(1,-0.15)(2,8.55)(4,13.95)(8,17.80)(16,21.60)(32,24.25)(64,26.15)(128,27.65)};
\nextgroupplot[title={7B, GSM8K}, ymin=-5, ymax=5]
\draw[cbase,dashed,line width=0.5pt] (axis cs:0.85,0) -- (axis cs:150,0);
\addplot[cthicket, mark=*, mark size=1.4pt, forget plot] coordinates {(1,0.83)(2,1.44)(4,2.27)(8,2.05)(16,1.97)(32,2.20)(64,2.43)(128,2.27)};
\addplot[clogit, mark=square*, mark size=1.3pt, forget plot] coordinates {(1,-1.21)(2,-0.83)(4,2.50)(8,3.49)(16,4.02)(32,4.47)(64,4.47)(128,4.40)};
\addplot[cttmv, mark=triangle*, mark size=1.6pt, forget plot] coordinates {(1,-0.91)(2,-0.45)(4,0.53)(8,1.74)(16,1.82)(32,1.90)(64,1.67)(128,1.97)};
\nextgroupplot[title={7B, MATH-500}, ymin=-10, ymax=25]
\draw[cbase,dashed,line width=0.5pt] (axis cs:0.85,0) -- (axis cs:150,0);
\addplot[cthicket, mark=*, mark size=1.4pt, forget plot] coordinates {(1,0.67)(2,1.00)(4,8.67)(8,13.00)(16,15.00)(32,17.67)(64,17.34)(128,17.00)};
\addplot[clogit, mark=square*, mark size=1.3pt, forget plot] coordinates {(1,9.00)(2,9.67)(4,13.67)(8,14.67)(16,16.67)(32,18.00)(64,18.34)(128,18.67)};
\addplot[cttmv, mark=triangle*, mark size=1.6pt, forget plot] coordinates {(1,-7.66)(2,-5.33)(4,3.67)(8,9.34)(16,11.67)(32,13.67)(64,14.34)(128,14.67)};
\end{groupplot}
\end{tikzpicture}

%% file: figures/fig_complementarity_bars.tex
\centering
\begin{tikzpicture}
\begin{axis}[paperaxis, width=\linewidth, height=5.4cm,
  ybar, bar width=3.1pt, enlarge x limits=0.035,
  symbolic x coords={0.5B Countdown,0.5B GSM8K,0.5B MATH-500,1.5B Countdown,1.5B GSM8K,1.5B MATH-500,3B Countdown,3B GSM8K,3B MATH-500,7B Countdown,7B GSM8K,7B MATH-500},
  xtick=data, x tick label style={rotate=32, anchor=east, font=\scriptsize},
  ymin=0, ymax=46, ylabel={peak gain over base (pts)},
  legend columns=4, legend style={at={(0.5,1.03)}, anchor=south, /tikz/every even column/.append style={column sep=7pt}}]
\addplot[fill=cthicket, draw=cthicket!55!black] coordinates {(0.5B Countdown,12.45)(0.5B GSM8K,11.67)(0.5B MATH-500,10.33)(1.5B Countdown,26.15)(1.5B GSM8K,16.68)(1.5B MATH-500,11.00)(3B Countdown,44.00)(3B GSM8K,6.52)(3B MATH-500,9.33)(7B Countdown,27.35)(7B GSM8K,2.43)(7B MATH-500,17.67)}; \addlegendentry{full-weight thicket}
\addplot[fill=clogit, draw=clogit!55!black] coordinates {(0.5B Countdown,12.60)(0.5B GSM8K,13.04)(0.5B MATH-500,13.00)(1.5B Countdown,27.95)(1.5B GSM8K,22.07)(1.5B MATH-500,13.34)(3B Countdown,44.25)(3B GSM8K,8.34)(3B MATH-500,10.00)(7B Countdown,27.25)(7B GSM8K,4.47)(7B MATH-500,18.67)}; \addlegendentry{logit bias}
\addplot[fill=csoup, draw=csoup!55!black] coordinates {(0.5B Countdown,11.10)(0.5B GSM8K,12.73)(0.5B MATH-500,11.67)(1.5B Countdown,17.10)(1.5B GSM8K,21.54)(1.5B MATH-500,12.34)(3B Countdown,44.25)(3B GSM8K,8.11)(3B MATH-500,11.33)(7B Countdown,26.55)(7B GSM8K,4.70)(7B MATH-500,19.67)}; \addlegendentry{thicket $\circ$ logit}
\addplot[fill=cextra, draw=cextra!55!black] coordinates {(0.5B Countdown,13.05)(0.5B GSM8K,13.42)(0.5B MATH-500,15.00)(1.5B Countdown,26.20)(1.5B GSM8K,21.92)(1.5B MATH-500,12.67)(3B Countdown,44.10)(3B GSM8K,8.19)(3B MATH-500,11.66)(7B Countdown,25.70)(7B GSM8K,4.85)(7B MATH-500,19.34)}; \addlegendentry{logit $\circ$ logit (control)}
\end{axis}
\end{tikzpicture}

%% file: figures/fig_complementarity_scatter.tex
\centering
\begin{minipage}[t]{0.49\linewidth}
\centering
\textbf{\footnotesize (a) stacked vs.\ best single method}\\[2pt]
\begin{tikzpicture}
\begin{axis}[paperaxis, width=\linewidth, height=5.6cm,
  xlabel={best single method peak (\%)}, ylabel={thicket $\circ$ logit peak (\%)},
  xmin=0, xmax=100, ymin=0, ymax=100, xtick={0,25,50,75,100}, ytick={0,25,50,75,100},
  legend style={at={(0.97,0.03)}, anchor=south east, font=\tiny}]
\addplot[gray, dashed, line width=0.5pt, domain=0:100, forget plot]{x};
\addplot[only marks, mark=triangle*, mark size=1.7pt, clogit] coordinates {(12.90,11.40)(57.32,57.01)(42.00,40.67)}; \addlegendentry{0.5B}
\addplot[only marks, mark=square*, mark size=1.5pt, cttmv] coordinates {(40.45,29.60)(82.49,81.96)(61.67,60.67)}; \addlegendentry{1.5B}
\addplot[only marks, mark=diamond*, mark size=1.7pt, cextra] coordinates {(57.90,57.90)(89.31,89.08)(72.67,74.00)}; \addlegendentry{3B}
\addplot[only marks, mark=*, mark size=1.5pt, cthicket] coordinates {(93.95,93.15)(91.96,92.19)(82.00,83.00)}; \addlegendentry{7B}
\end{axis}
\end{tikzpicture}
\end{minipage}
\hfill
\begin{minipage}[t]{0.49\linewidth}
\centering
\textbf{\footnotesize (b) stacked vs.\ same-mechanism control}\\[2pt]
\begin{tikzpicture}
\begin{axis}[paperaxis, width=\linewidth, height=5.6cm,
  xlabel={logit $\circ$ logit control peak (\%)}, ylabel={thicket $\circ$ logit peak (\%)},
  xmin=0, xmax=100, ymin=0, ymax=100, xtick={0,25,50,75,100}, ytick={0,25,50,75,100},
  legend style={at={(0.97,0.03)}, anchor=south east, font=\tiny}]
\addplot[gray, dashed, line width=0.5pt, domain=0:100, forget plot]{x};
\addplot[only marks, mark=triangle*, mark size=1.7pt, clogit] coordinates {(13.35,11.40)(57.70,57.01)(44.00,40.67)}; \addlegendentry{0.5B}
\addplot[only marks, mark=square*, mark size=1.5pt, cttmv] coordinates {(38.70,29.60)(82.34,81.96)(61.00,60.67)}; \addlegendentry{1.5B}
\addplot[only marks, mark=diamond*, mark size=1.7pt, cextra] coordinates {(57.75,57.90)(89.16,89.08)(74.33,74.00)}; \addlegendentry{3B}
\addplot[only marks, mark=*, mark size=1.5pt, cthicket] coordinates {(92.30,93.15)(92.34,92.19)(82.67,83.00)}; \addlegendentry{7B}
\end{axis}
\end{tikzpicture}
\end{minipage}